%% file: main.tex
\PassOptionsToPackage{table}{xcolor}
\documentclass{reweave}
\input{preamble}

\makeatletter
\providecommand\@LN[2]{}
\providecommand\@LN@col[1]{}
\makeatother

\DeclareRobustCommand{\reweavetitlemark}{\raisebox{-0.08em}{\includegraphics[height=.86em]{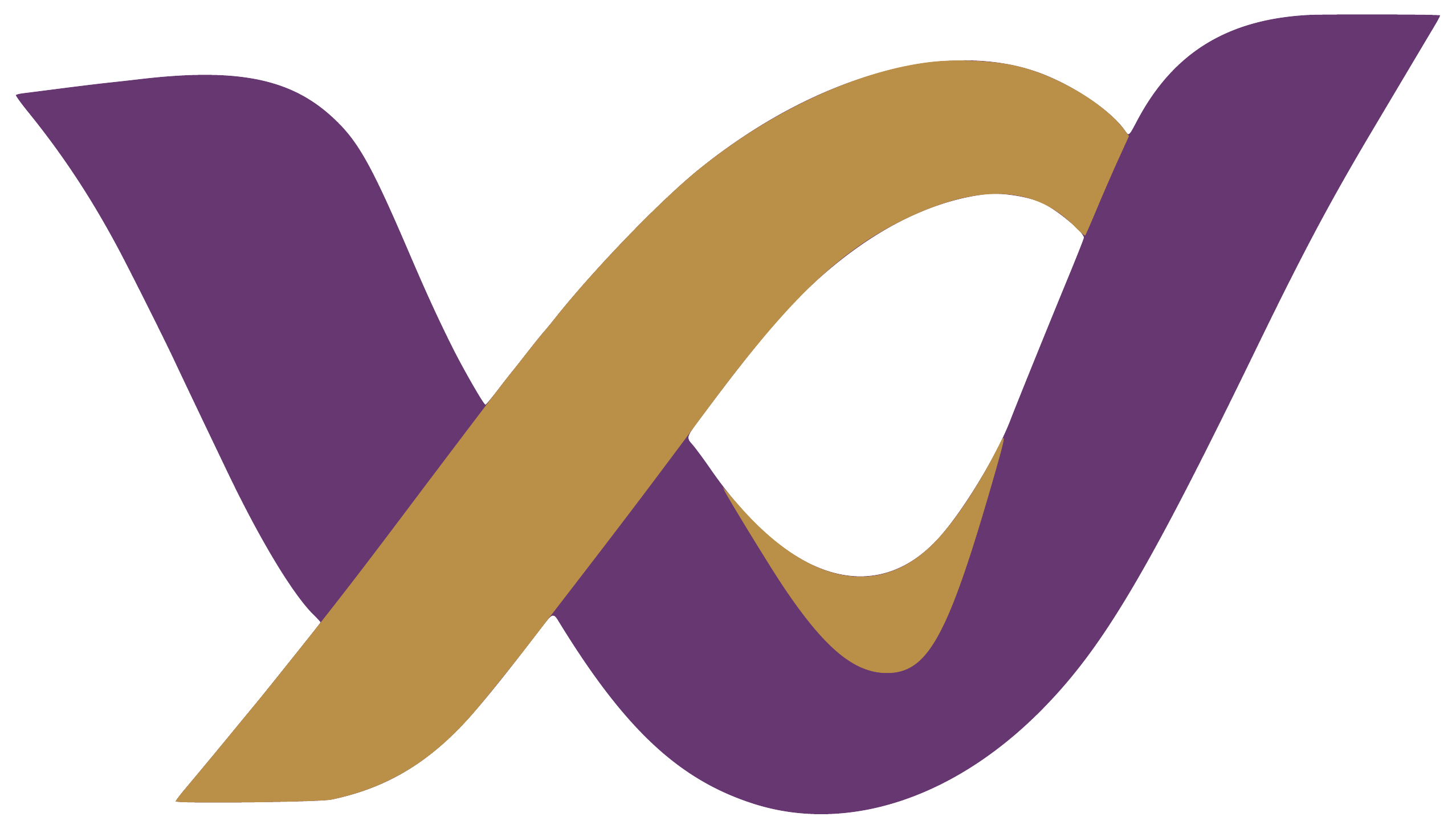}}\hspace{.23em}}
\title{\texorpdfstring{\reweavetitlemark}{}\papermaintitle\space\papersubtitle}
\coverlogo{%
  \includegraphics[height=0.70cm]{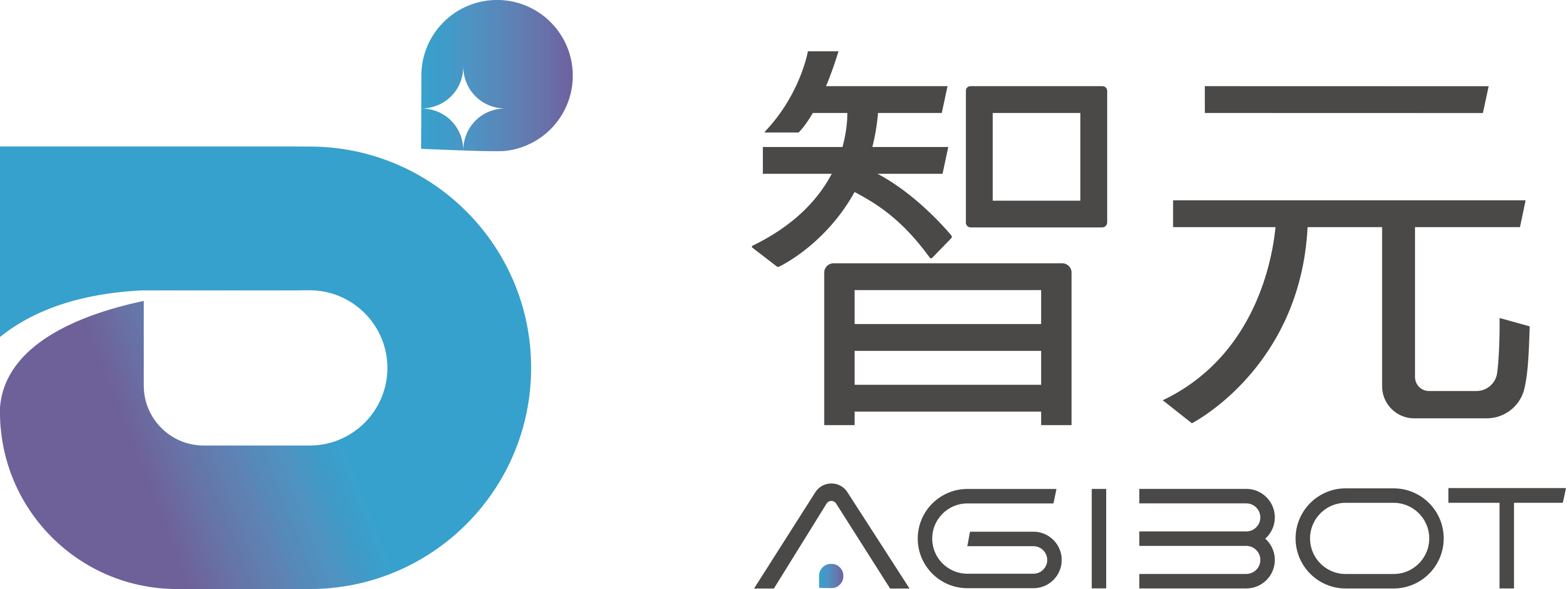}%
  \hspace{0.34cm}%
  \raisebox{0.01cm}{\includegraphics[height=0.72cm]{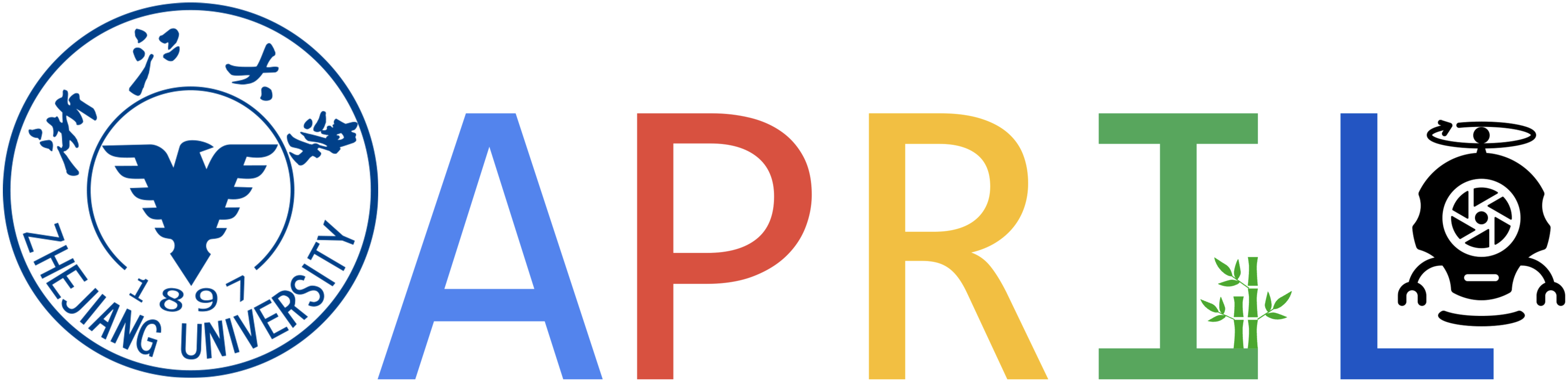}}}
\author[1]{Yicheng~Xu}
\author[1]{Wencan~Jiang}
\author[2,4]{Zihao~Zhu}
\author[3]{Keru~Zhou}
\author[4]{Chang~Zhou}
\authorbreak
\author[4]{Li~Liu}
\author[4]{Maoqing~Yao}
\author[1]{Yong~Liu}
\author[1,\dagger]{Jiangning~Zhang}
\affiliation[1]{Zhejiang University}
\affiliation[2]{Stanford University}
\affiliation[3]{Tsinghua University}
\affiliation[4]{AGIBOT Innovation}
\contribution[\dagger]{Corresponding author}
\reweaveabstract{\input{sections/00_abstract}}
\coverdate{November 24, 2026}
\coversourcecode{https://github.com/xuyicheng-zju/ReWeave}
\coverproject{https://xuyicheng-zju.github.io/ReWeave/}
\covercorrespondence{186368@zju.edu.cn}
\reweaverunning{ReWeave}
\hypersetup{pdftitle={\papermaintitle\space\papersubtitle},
  pdfauthor={Yicheng Xu, Wencan Jiang, Zihao Zhu, Keru Zhou, Chang Zhou,
    Li Liu, Maoqing Yao, Yong Liu, Jiangning Zhang}}

\begin{document}
\maketitle
\vspace{8pt}
\noindent
\begin{minipage}{\textwidth}
  \captionsetup{type=figure}
  \input{figures/overview}
\end{minipage}

\twocolumn
\flushbottom
\input{sections/01_introduction}
\input{sections/02_related_work}
\input{sections/04_methodology}
\input{sections/05_experiments}
\input{sections/06_conclusion}

\bibliographystyle{reweave}
\bibliography{references}

\onecolumn
\raggedbottom
\reweavestamp{Supplementary material}
\pagestyle{reweavesupplement}
\appendix
\input{sections/07_appendix}
\end{document}

%% file: preamble.tex
\usepackage{hyperref}
\hypersetup{colorlinks,breaklinks,linkcolor=echoaccent,citecolor=echoaccent,urlcolor=echoaccent}

\usepackage{graphicx}
\usepackage{array}
\usepackage{multirow}
\usepackage{tabularx}
\usepackage{longtable}
\usepackage{pifont}
\usepackage{algorithm}
\usepackage[noend]{algpseudocode}
\usepackage{float}
\usepackage{placeins}
\usepackage{colortbl}
\graphicspath{{figures/}}

\input{macros}

\newlength{\reweavealgorithmruleheight}
\makeatletter
\renewcommand\floatc@ruled[2]{{\@fs@cfont\textcolor{echoaccent}{#1}\hspace{0.45em}}{\bfseries\color{echoink}#2}\par}
\renewcommand\fs@ruled{%
  \def\@fs@cfont{\bfseries}%
  \let\@fs@capt\floatc@ruled
  \def\@fs@pre{{\color{echoaccent}\hrule height\reweavealgorithmruleheight depth 0pt}\kern 3pt}%
  \def\@fs@post{\kern 3pt{\color{echoaccent}\hrule height\reweavealgorithmruleheight depth 0pt}\relax}%
  \def\@fs@mid{\kern 3pt{\color{echoborder}\hrule height 0.5pt depth 0pt}\kern 3pt}%
  \let\@fs@iftopcapt\iftrue}
\makeatother
\floatstyle{ruled}
\restylefloat{algorithm}

\makeatletter
\providecommand{\theHALG@line}{\arabic{algorithm}.\arabic{ALG@line}}
\renewcommand{\theHALG@line}{\arabic{algorithm}.\arabic{ALG@line}}
\makeatother

%% file: macros.tex
\newcommand{\echol}[1]{\textcolor{echoaccent}{#1}}
\newcommand{\papermaintitle}{\echol{ReWeave:} Emergent Physical Task Topologies}
\newcommand{\papersubtitle}{for Embodied Harness Optimization and Cross-Task Meta-Distillation}

\newif\ifpaperbibliography
\paperbibliographytrue

\newif\ifpapercoloredlinks
\papercoloredlinkstrue

\definecolor{tarsacpurple}{HTML}{8557A0}
\definecolor{tableaccent}{HTML}{EFE8F5}
\definecolor{deltaselected}{HTML}{F5C75C}
\definecolor{tablegray}{HTML}{F1EEF6}
\definecolor{panelgray}{RGB}{240,240,240}
\colorlet{echoaccent}{tarsacpurple}
\definecolor{echolink}{HTML}{6B2FB5}

\newcommand{\bestval}[1]{\textbf{#1}}
\newcommand{\runnerupval}[1]{\underline{#1}}
\newcommand{\scorepair}[2]{\mbox{\makebox[2.3em][c]{#1}\kern0.8pt/\kern0.8pt\makebox[2.3em][c]{#2}}}
\newcommand{\deltagain}[1]{\textcolor{ablationgain}{\textbf{+#1}}}
\newcommand{\deltapair}[2]{\mbox{\scriptsize\deltagain{#1}\kern0.4pt/\kern0.4pt\deltagain{#2}}}
\newcommand{\scorepairgain}[3]{\mbox{\makebox[1.7em][c]{#1}\kern0.8pt/\kern0.8pt\makebox[1.7em][c]{#2}\textsuperscript{\fontsize{5.5}{6}\selectfont\textcolor{ablationgain}{+#3}}}}
\newcommand{\methodlabel}[1]{\textcolor{echoaccent}{\textbf{#1}}}
\newcommand{\challenge}[3]{%
  \par\smallskip\vspace{2pt}\noindent
  \hypertarget{challenge:#1}{}%
  \textcolor{echoaccent}{\textbf{[Challenge~#1]\enspace #2}%
    \if\relax\detokenize{#3}\relax\else\enspace\textbf{[#3]}\fi.}\enspace}
\newcommand{\challengeref}[1]{\hyperlink{challenge:#1}{\textcolor{echoaccent}{\textbf{Challenge~#1}}}}

\newcommand{\finding}[2]{\textbf{\textit{(#1) #2}}\enspace}
\definecolor{ablationdrop}{HTML}{A64B32}
\definecolor{ablationgain}{HTML}{16704A}
\newcommand{\yes}{\textcolor{ablationgain}{\ding{51}}}
\newcommand{\no}{\textcolor{ablationdrop}{\ding{55}}}

\newcommand{\gainval}[2]{\mbox{#1\textsuperscript{\fontsize{5.5}{6}\selectfont\textcolor{ablationgain}{+#2}}}}
\newcolumntype{L}[1]{W{l}{#1}}
\newcolumntype{C}[1]{W{c}{#1}}
\newcolumntype{R}[1]{W{r}{#1}}
\newcommand{\tablehead}[1]{\begin{tabular}[c]{@{}c@{}}#1\end{tabular}}
\newcommand{\tabgroup}[2]{\rowcolor{tablegray}\multicolumn{#1}{l}{\rule{0pt}{2.0ex}\textbf{#2}}\\}

\newcommand{\keyobservation}[2][K{\footnotesize EY} O{\footnotesize BSERVATION}]{%
  \par\smallskip\noindent
  \begingroup
  \setlength{\fboxsep}{5.5pt}%
  \colorbox{tableaccent!55}{%
    \begin{tabular}{@{\color{echoaccent!85}\vrule width 1.6pt\hspace{0.55em}}p{\dimexpr\linewidth-2\fboxsep-1.6pt-0.55em\relax}@{}}
      \textcolor{echoaccent}{\textbf{#1.}}\enspace\textit{#2}\par
    \end{tabular}%
  }%
  \endgroup
  \par\smallskip
}
\newcommand{\researchquestion}[1]{\keyobservation[Research question]{#1}}

\newcommand{\methodinsight}[2]{%
  \par\smallskip\noindent
  \begin{tabular}{@{\color{echoaccent!85}\vrule width 1.2pt\hspace{0.6em}}p{\dimexpr\linewidth-1.2pt-0.6em\relax}@{}}
    \setlength{\emergencystretch}{3em}\textcolor{echoaccent}{\textbf{[Insight~#1]}}\enspace\textit{#2}\par
  \end{tabular}%
  \par\smallskip
}

\newcommand{\sys}{ReWeave}
\newcommand{\methodname}{\sys}

\newcommand{\algphase}[1]{%
  \vspace{2pt}%
  \Statex \textcolor{echoaccent}{\textbf{#1}}%
}
\newcommand{\algcmt}[1]{\hfill\textcolor{black!55}{\(\triangleright\)~\textit{#1}}}

%% file: sections/00_abstract.tex
Long-horizon embodied tasks involve physical dependencies that may only become clear through interaction. As task understanding deepens, past trials can reveal new information for adaptation and transfer. However, coarse progress feedback cannot identify which physical preconditions held, leaving search unguided, while task-specific optimization easily overfits to individual scenes, making cross-task transfer brittle. We introduce \methodname{}, a framework that jointly optimizes an embodied agent harness and an emergent physical task topology from interaction without updating policy weights. Keeping execution policies and perception frozen, \methodname{} co-optimizes coupled context, workflows, and control loops. As new evidence revises the topology, \methodname{} reprojects past trials onto the updated task understanding, exposing retained progress, regressions, and bottlenecks that guide parent selection and search width. Across tasks, matching verified local mechanisms to unmet preconditions consolidates $72$ specialist units into $44$ reusable groups ($0.34\times$ full concatenation) within a compact shared harness capable of continual learning. \methodname{} improves held-out success from $89.00\%$ to $98.75\%$ on LIBERO-PRO and from $25.50\%$ to $58.25\%$ on RoboDojo, while uncovering retained progress in $89\%$ of $159$ failed RoboDojo development trials and translating stronger offline reasoning into higher online performance. On eight unseen LIBERO-PRO tasks, the shared harness reaches $51.3\%$ success versus $41.3\%$ for direct distillation, rising to $75.0\%$ after continual learning.

%% file: figures/overview.tex
\centering
\captionsetup{hypcap=false}
\includegraphics[width=\textwidth]{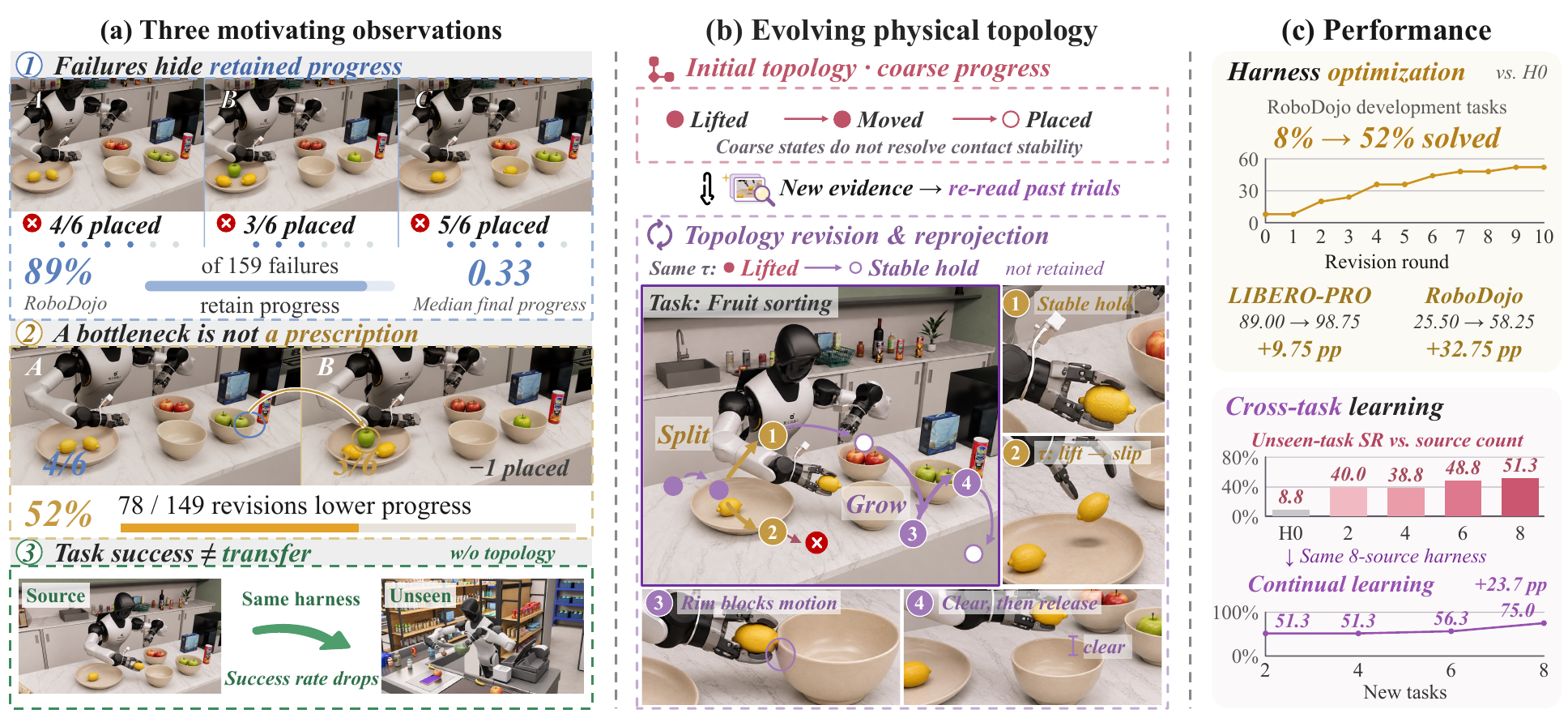}
\caption{\textbf{From coarse progress to emergent physical topology.} \textbf{\textit{(a)}} Retained progress and revision regressions motivate \methodname{}; illustrative G2 scenes depict the source-to-unseen transfer gap. \textbf{\textit{(b)}} Illustrative G2 evidence separates stable hold from slip, adds rim clearance as a prerequisite, and reinterprets the same trial. \textbf{\textit{(c)}} Harness optimization, cross-task transfer, and continual learning.}
\label{fig:teaser}

%% file: sections/01_introduction.tex
\section{Introduction}
\label{sec:introduction}

Long-horizon manipulation depends on establishing physical conditions needed by downstream actions~\cite{liang2023code,huang2022inner}. Plausible high-level plans often fail because key physical requirements, such as contact stability, clearance margins, and assembly tolerances, emerge only during execution. These dynamics cannot be anticipated \textit{a priori}, making physical trial-and-error necessary yet costly. Maximizing what each trial teaches is thus vital for adaptation and reuse.

Recent work repairs programs and evolves skills from execution feedback~\cite{wang2026self,ding2026zetta,lu2026aspire}, distills playbooks~\cite{kim2026recursive}, or organizes repairs with workflow graphs~\cite{wang2026harnesspai}. These advances leave a distinct question: when new evidence changes our understanding of a physical prerequisite, how should already collected trials be reassessed? Keeping their interpretation fixed can hide partial progress behind failure labels. Likewise, storing a successful routine does not identify which local mechanism another task needs. Revisable dependencies connect retrospective credit, harness revision, and condition-based transfer (\hyperref[fig:teaser]{Fig.~\ref*{fig:teaser}}).

\researchquestion{How can we continually revise physical task understanding through interaction to reinterpret experience, improve agent harnesses, and transfer mechanisms across tasks?}

\challenge{1}{Nominal task plans mask latent physical bottlenecks}{\hyperref[sec:emergent-topology]{Sec.~\ref*{sec:emergent-topology}}}
Execution can expose missing physical requirements. Revised understanding calls for re-reading past trials to assess retained progress.

\challenge{2}{Diagnosing a bottleneck does not reveal a remedy}{\hyperref[sec:harness-optimization]{Sec.~\ref*{sec:harness-optimization}}}
Context, action routines, and feedback loops are coupled. Several edits can address one failure, but each needs physical validation.

\challenge{3}{Task-specific success does not guarantee transfer}{\hyperref[sec:shared-distillation]{Sec.~\ref*{sec:shared-distillation}}}
Specialists depend on their scenes and objects. New tasks may need one local mechanism or parts from several specialists, each with prerequisites.

\setcounter{dbltopnumber}{1}
\begin{figure*}[t]
  \centering
  \includegraphics[width=\textwidth]{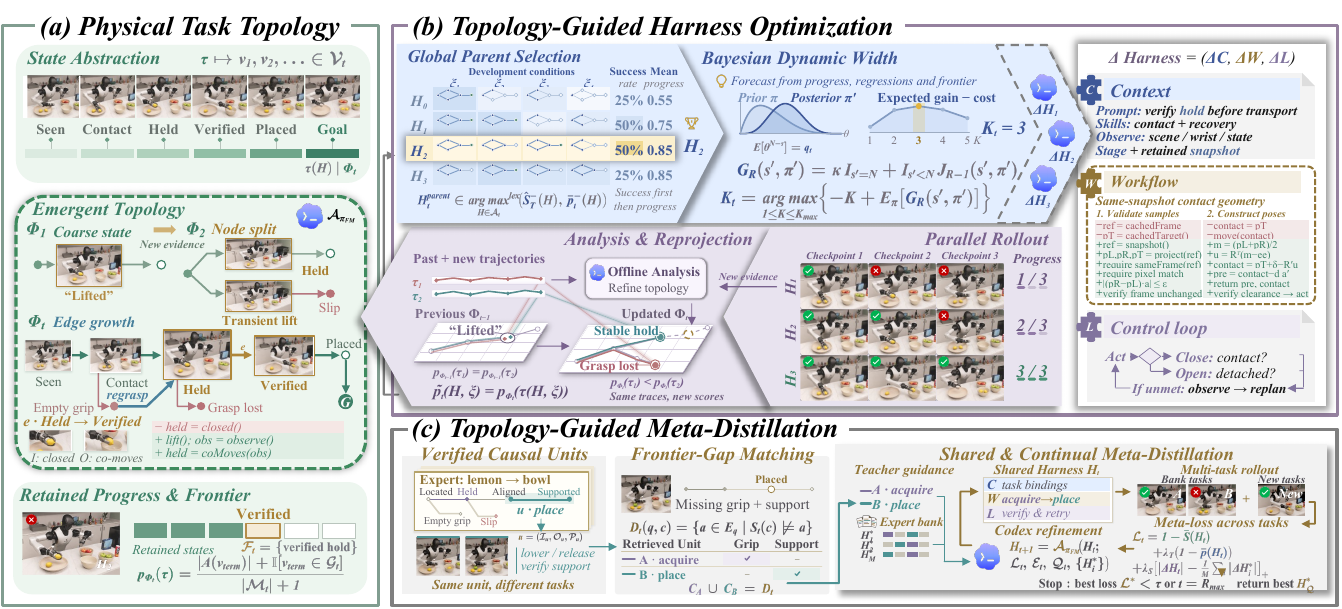}
  \caption{\textbf{Overview of \methodname{}.} \textbf{\textit{(a)}} Physical trials induce and refine a task topology to re-read progress. \textbf{\textit{(b)}} Topology steers parent selection and search width. \textbf{\textit{(c)}} Causal units guide specialist consolidation and continual learning.}
  \label{fig:method}
\end{figure*}

We introduce \methodname{}, which uses an \textbf{\textit{emergent physical task topology}} to make changes in task understanding explicit (\hyperref[fig:method]{Fig.~\ref*{fig:method}}). It links physical dependencies and alternative routes to evidence and harness mechanisms. Beyond using old traces to propose repairs, topology revision changes how the same archived trials are credited and compared (\challengeref{1}). Reprojected progress and bottlenecks then guide harness search (\challengeref{2}). Across tasks, matching verified transition effects to unmet conditions identifies which specialist mechanisms to combine in a shared harness for transfer and continual learning (\challengeref{3}).

On LIBERO-PRO and RoboDojo, \methodname{} achieves strong performance: $98.75\%$ and $58.25\%$ SR, gains of $9.75$ and $32.75$ points over $H_0$, respectively. Re-reading 159 failed RoboDojo development trials on their final topologies finds retained progress in $89\%$ (median $0.33$; \hyperref[fig:topology-read]{Fig.~\ref*{fig:topology-read}}). On unseen tasks, direct distillation reaches $41.3\%$ SR and the shared harness reaches $51.3\%$.

\begin{samepage}
Our contributions are threefold:
\begin{itemize}
  \item \textbf{Emergent Physical Task Topology.} We introduce an emergent representation of task understanding that connects physical dependencies, alternative routes, execution evidence, and reusable harness mechanisms.
  \item \textbf{Topology-Guided Harness Optimization.} We develop joint harness optimization guided by emergent task understanding and the physical cost of further search.
  \item \textbf{Topology-Guided Meta-Distillation.} We develop meta-distillation matching specialist mechanisms to unmet conditions for cross-task consolidation and transfer.
\end{itemize}
\end{samepage}

%% file: sections/02_related_work.tex
\section{Related Work}
\label{sec:related-work}

\subsection{Embodied Agent Harnesses}
Language-guided robots use skills, policies, feedback~\cite{ahn2022can,liang2023code,huang2022inner}, and scene representations of preconditions~\cite{rana2023sayplan,han2024interpret}. Robot-use agents expose perception and action interfaces~\cite{ora2026code,chu2026openrua,gao2026hexashell,li2026roboclaw}, while harnesses add memory, orchestration, and state tracking~\cite{zhang2026harness,huang2026roboharness,gu2026harnesswam,xu2026tarsphere}. These interfaces let an agent inspect the scene and act, while the harness organizes how that information and feedback enter execution. In multimodal in-context learning, UniICL~\cite{xu2026uniicl} organizes task-aligned demonstrations by capability. \methodname{} jointly adapts context, workflows, and control loops while preserving fixed embodiment interfaces.

\subsection{Embodied Harness Optimization}
Harness optimization refines skills, context, and routines from physical feedback~\cite{wang2026self,du2026vlm,elmaaroufi2026rho,lu2026aspire,ding2026zetta} or distills interventions~\cite{li2026embodiedevo,kim2026recursive}; ENPIRE adapts real-world policies~\cite{xiao2026enpire}. HarnessPAI can use a fixed workflow graph from expert video while optimizing its code and repair skills~\cite{wang2026harnesspai}.
\methodname{} represents evolving physical dependencies in a revisable topology. It re-scores archived trials to guide search and matches unmet conditions to reusable local mechanisms.

%% file: sections/04_methodology.tex
\section{Method}
\label{sec:method}
\begingroup
\setcounter{secnumdepth}{3}
\setlength{\abovedisplayskip}{3pt plus 2pt}
\setlength{\belowdisplayskip}{3pt plus 2pt}
\setlength{\abovedisplayshortskip}{2pt plus 2pt}
\setlength{\belowdisplayshortskip}{2pt plus 2pt}
\setlength{\jot}{2pt}

\subsection{Problem Formulation}
\label{sec:problem-formulation}

\noindent\textbf{Coupled {\boldmath$(C,W,L)$} Solution Space.} An inner agent based on a frozen foundation model $\pi_{\mathrm{FM}}$ acts as the policy; an outer agent revises its harness between episodes:
\begingroup
\setlength{\abovedisplayskip}{2pt}
\setlength{\belowdisplayskip}{2pt}
\setlength{\abovedisplayshortskip}{2pt}
\setlength{\belowdisplayshortskip}{2pt}
\begin{equation}
    H = (C,W,L) \in \mathcal{H}_{\mathrm{CWL}},
    \label{eq:harness-space}
\end{equation}
\endgroup
where context $C$ curates instructions, workflow $W$ defines action maneuvers, and control loop $L$ governs feedback timing. Isolated revisions leave bottlenecks unresolved; their resolution demands joint changes across coupled context, workflows, and feedback.

\noindent{\textbf{Harness-of-Agent Optimization.}} For a partially observed long-horizon task $T$, a rollout $\tau(H,\xi)$ with $\xi\sim\mathcal D_T$ receives an official success label $S_T(H,\xi)\in\{0,1\}$. The specialist objective is
\begingroup
\setlength{\abovedisplayskip}{0pt}
\setlength{\belowdisplayskip}{0pt}
\setlength{\abovedisplayshortskip}{0pt}
\setlength{\belowdisplayshortskip}{0pt}
\begin{equation}
 H_T^{\mathrm{opt}}\in\arg\max_{H}\ \mathbb E_{\xi\sim\mathcal D_T}[S_T(H,\xi)].
 \label{eq:harness-objective}
\end{equation}
\endgroup
With bounded development feedback, search returns a task-specific specialist \mbox{$H_T^*$}, and distillation consolidates a shared harness for transfer to unseen tasks (\hyperref[sec:shared-distillation]{Sec.~\ref*{sec:shared-distillation}}).

\subsection{ReWeave Framework}
\label{sec:reweave-framework}

\sys{} refines a physical task topology from trials to guide two optimization loops (\hyperref[fig:method]{Fig.~\ref*{fig:method}}). Within a task, reprojecting past trajectories guides harness search (\hyperref[alg:coevolution]{Algorithm~\ref*{alg:coevolution}}); across tasks, local transitions identify reusable mechanisms for consolidation and transfer (\hyperref[alg:distillation]{Algorithm~\ref*{alg:distillation}}).

\makeatletter
\begin{figure*}[t]
\let\paired@algorithmstyle\fs@ruled
\def\fs@ruled{%
  \paired@algorithmstyle
  \def\@fs@post{\kern3pt}%
}
\noindent
\begin{minipage}[t]{\columnwidth}
\vspace{0pt}
\begin{algorithm}[H]
\captionsetup{hypcap=false}
\caption{Topology-guided harness optimization}
\phantomsection
\label{alg:coevolution}
\footnotesize
\raggedright
\begin{algorithmic}[1]
\Require $\pi_{\mathrm{FM}}$, seed $H_0\in\mathcal H_{\mathrm{CWL}}$, trials $\Xi_T$, rounds $T_{\max}$.
\Ensure Specialist harness $H_T^*$, final topology $\Phi_{\mathrm{f}}$.
\State $\mathcal T_{\le0}\gets\operatorname{Rollout}(H_0,\Xi_T;\pi_{\mathrm{FM}})$, $\mathcal A_1\gets\{H_0\}$, $\Phi_0\gets\varnothing$
\For{round $t=1,\ldots,T_{\max}$ until $\Xi_T$ is solved}
\algphase{Stage 1: topology refinement and width allocation}
\State Induce $\Phi_t$ from $\mathcal T_{\le t-1}$ \algcmt{\hyperref[eq:task-topology]{Eq.~(\ref*{eq:task-topology})}}
\ForAll{$H\in\mathcal A_t$, $\xi\in\mathcal O_t(H)$}
\State Reproject progress, loss, and frontier \algcmt{\hyperref[eq:retained-progress]{Eqs.~(\ref*{eq:retained-progress})}-\hyperref[eq:reprojection-frontier]{(\ref*{eq:reprojection-frontier})}}
\EndFor
\State $H_t^{\mathrm{parent}}\gets\arg\max^{\mathrm{lex}}_{H\in\mathcal A_t}(\hat S_T^-,\bar p_t^-)$ \algcmt{\hyperref[eq:parent-selection]{Eq.~(\ref*{eq:parent-selection})}}
\State Forecast $q_t$ and set width $K_t$ \algcmt{\hyperref[eq:search-width]{Eq.~(\ref*{eq:search-width})}}
\algphase{Stage 2: evidence-guided synthesis and rollout}
\For{candidate $k=1,\ldots,K_t$ \textbf{in parallel}}
\State Collect witnesses $\mathcal Z_t(\mathcal U_t)$ of unresolved transitions
\State $H_{t,k}\gets H_t^{\mathrm{parent}}\oplus\Delta H_{t,k}$ \algcmt{\hyperref[eq:evidence-guided-synthesis]{Eq.~(\ref*{eq:evidence-guided-synthesis})}}
\If{$H_{t,k}$ satisfies execution limits}
\State Add $\operatorname{Rollout}(H_{t,k},\Xi_T;\pi_{\mathrm{FM}})$ to $\mathcal T_{\le t}$, $H_{t,k}$ to $\mathcal A_{t+1}$
\EndIf
\EndFor
\EndFor
\State Induce final $\Phi_{\mathrm f}$ from $\mathcal T_{\le T_{\max}}$
\State \Return best archived $H$ by \hyperref[eq:parent-selection]{Eq.~(\ref*{eq:parent-selection})}, $\Phi_{\mathrm f}$
\end{algorithmic}
\end{algorithm}
\unskip
\end{minipage}\hfill
\begin{minipage}[t]{\columnwidth}
\vspace{0pt}
\begin{algorithm}[H]
\captionsetup{hypcap=false}
\caption{Topology-guided meta-distillation}
\phantomsection
\label{alg:distillation}
\footnotesize
\raggedright
\begin{algorithmic}[1]
\Require $H_{\mathrm{init}}$, tasks $\mathcal T$, query tasks $\mathcal T_{\mathrm q}$, bank $\mathcal B$, panel $\Xi_{\mathcal T}$, threshold $\tau$, rounds $R_{\max}$.
\Ensure Shared harness $H_{\mathrm{sh}}^*\in\mathcal H_{\mathrm{CWL}}$.
\State $\mathcal U^+\gets$ verified units of $\mathcal B$; $H_1\gets H_{\mathrm{init}}$; $\mathcal L^*\gets\infty$
\For{round $t=1,\ldots,R_{\max}$ until $\mathcal L^*<\tau$}
\State $\mathcal E_t\gets\operatorname{Rollout}(H_t,\Xi_{\mathcal T};\pi_{\mathrm{FM}})$
\State $\mathcal L_t\gets\mathcal L_{\mathrm{meta}}(H_t;\mathcal E_t)$ \algcmt{\hyperref[eq:shared-distillation]{Eq.~(\ref*{eq:shared-distillation})}}
\State \textbf{if} $\mathcal L_t<\mathcal L^*$ \textbf{then} $(H_{\mathrm{sh}}^*,\mathcal L^*)\gets(H_t,\mathcal L_t)$
\algphase{Stage 1: frontier gaps and unit retrieval}
\State Refresh new-task topologies; $\mathcal Q_t\gets\varnothing$
\ForAll{failed trial $c$ on $\mathcal T_{\mathrm q}$}
\State $F_t(c)\gets\operatorname{Frontier}(\Phi_i,c)$ \algcmt{\hyperref[eq:reprojection-frontier]{Eq.~(\ref*{eq:reprojection-frontier})}}
\ForAll{transition $q$ entering $F_t(c)$}
\State Gap $D_t(q,c)$, bindings $R_t(q,c)$ \algcmt{\hyperref[eq:dependency-gap]{Eqs.~(\ref*{eq:dependency-gap})}-\hyperref[eq:topology-alignment]{(\ref*{eq:topology-alignment})}}
\State $\mathcal Q_t\gets\mathcal Q_t\cup\{(c,q,D_t(q,c),R_t(q,c))\}$
\EndFor
\EndFor
\algphase{Stage 2: loss-guided meta-distillation}
\State $H_{t+1}\gets\mathcal A_{\pi_{\mathrm{FM}}}(H_t;\mathcal L_t,\mathcal E_t,\mathcal Q_t,\mathcal B)$ \algcmt{\hyperref[eq:harness-update]{Eq.~(\ref*{eq:harness-update})}}
\EndFor
\State \Return $H_{\mathrm{sh}}^*$
\end{algorithmic}
\end{algorithm}
\unskip
\end{minipage}\par
\nointerlineskip
\hbox to\textwidth{%
  \color{echoaccent}%
  \vrule width\columnwidth height\reweavealgorithmruleheight depth0pt\hfil
  \vrule width\columnwidth height\reweavealgorithmruleheight depth0pt
}
\end{figure*}
\makeatother

\subsubsection{Emergent Physical Task Topology}
\label{sec:emergent-topology}

\methodinsight{1}{\textcolor{echoaccent}{\textbf{How to interpret experience?}} A physical topology binds transitions to conditions, witnesses, and code, giving past trials a common coordinate system. When revised, old rollouts are re-projected on this updated metric, revealing progress their binary labels hid.}

\noindent\textbf{Evidence-Grounded Physical Topology.}
After each round, the outer agent reads the task and recorded observations and actions $\mathcal{T}_{\le t-1}$ and induces a task topology:
\begin{equation}
    \Phi_t = (\mathcal V_t,\mathcal E_t,\mathcal P_t), \qquad
    \mathcal V_t = \mathcal M_t\cup\mathcal B_t\cup\mathcal G_t,
\label{eq:task-topology}
\end{equation}
where $\mathcal M_t$, $\mathcal B_t$, and $\mathcal G_t$ are required, optional, and goal states. Required states follow prerequisite precedence $\prec_{\Phi_t}$. Each edge $e:u\to v$ binds entry/exit conditions $(\mathcal I_e,\mathcal O_e)$ to witnesses $\mathcal Z_e$ and code $\mathcal P_e\subseteq\mathcal P_t$. New evidence can split states, change which are required, and add paths. All archived trials use a \textbf{\textit{shared, revisable coordinate scale}}: they are re-segmented and reprojected onto the updated topology.
The outer agent revises the graph and archived state intervals in JSON. Host checks coverage and evidence links, then computes progress and frontiers (Supp.~\ref{sec:supp-method-implementation}).

\noindent\textbf{Topological Progress and Regression Decomposition.}
Let $A_{\Phi_t}(v)=\{m\in\mathcal M_t:m\preceq_{\Phi_t}v\}$ collect required milestones retained at node $v$. For trajectory $\tau$, write $A_{\mathrm{end}}$ for this set at termination and $A_{\mathrm{ever}}$ for its union over visited states. Retained progress and regression loss are
\begin{equation}
    p_{\Phi_t}(\tau)=\frac{|A_{\mathrm{end}}|+\mathbf 1_g}{|\mathcal M_t|+1},\;
    \ell_{\Phi_t}(\tau)=\frac{|A_{\mathrm{ever}}\setminus A_{\mathrm{end}}|}{|\mathcal M_t|+1}.
\label{eq:retained-progress}
\end{equation}
Here $\mathbf 1_g$ marks a terminal goal state. A dropped grasp incurs regression ($\ell_{\Phi_t}>0$); failing to grasp does not. Both are official failures, but the former proves the transition is attainable and lacks only retention.

\noindent\textbf{Reprojection and Frontier Localization.}
Each revision re-scores an archived rollout as $\tilde p_t(H,\xi)=p_{\Phi_t}(\tau(H,\xi))$. Its frontier contains unmet milestones whose strict required prerequisites $\operatorname{pre}_t(m)$ remain retained:
\nopagebreak
\begin{equation}
    \mathcal F_t=\{m\in\mathcal M_t\setminus A_{\mathrm{end}}:\operatorname{pre}_t(m)\subseteq A_{\mathrm{end}}\}.
\label{eq:reprojection-frontier}
\end{equation}
When the topology separates a stable grasp from a transient lift, trials once scored alike part ways, exposing hidden bottlenecks. The frontier marks what a trial can pursue next; its bottlenecks $\mathcal U_t(\xi)$, attempted but unrealized transitions, show where the harness fails to advance toward the goal.

\subsubsection{Topology-Guided Harness Optimization}
\label{sec:harness-optimization}
\label{sec:cwl-space}

\methodinsight{2}{\textcolor{echoaccent}{\textbf{How does better task understanding improve execution?}} On the emergent topology, the archive reveals which harness advanced furthest and what blocked it. This grounded assessment steers where search branches, what to revise, and how to allocate physical trials.}

\noindent\textbf{Topology-Guided Parent Selection.}
Search continues from the archived harness that solves the most development trials; among equals, re-read progress picks the one that has gone furthest on the current topology:
\nopagebreak
\begin{equation}
    H_t^{\mathrm{parent}}\in\arg\max_{H\in\mathcal A_t}^{\mathrm{lex}}\big(\hat S_T^-(H),\bar p_t^-(H)\big).
\label{eq:parent-selection}
\end{equation}
Here $\mathcal A_t$ is the harness archive and $\hat S_T^-$ is the confirmed success rate over $N$ development trials.

Supported mean progress over known outcomes $\mathcal O_t(H)$ is
\nopagebreak
\[
    \bar p_t^-(H)=\frac{1}{N}\sum_{\xi\in\mathcal O_t(H)}\tilde p_t(H,\xi).
\]
Reprojection restores an older harness when revised topologies credit its progress.

\noindent\textbf{Bayesian Cost-Aware Search Width.}
Under parallel rollout budgets, dynamic width trades extra rollouts for fewer sequential rounds. Given success probability $\theta$, candidate outcomes on $N-s$ unsolved conditions follow independent Bernoulli trials. From re-read progress and frontiers, the outer agent forecasts completion probability $q_t$ and sets a Beta prior with mean $m_t$ and strength $n_0$ (Supp.~\ref{sec:supp-method-implementation}):
\nopagebreak
\begin{equation}
    \theta \sim \mathrm{Beta}\big(n_0m_t,\,n_0(1-m_t)\big),\quad \mathbb E\big[\theta^{N-s}\big] = q_t.
\label{eq:width-prior}
\end{equation}
With $R$ rounds left, $K\le K_{\max}$ candidates cost $K$ evaluations and earn $\kappa$ upon completion; otherwise their outcomes update posterior $\pi=(\alpha,\beta)$. The recursion values information from failed rounds and yields optimal width $K_t$:
\nopagebreak
\begin{equation}
\begin{aligned}
    J_R(s,\pi) &= \max_{K}\Big\{-K +\mathbb E\big[\kappa\,\mathbb I_{s'=N} \\
    &\qquad +\mathbb I_{s'<N}\,J_{R-1}(s',\pi')\big]\Big\}.
\end{aligned}
\label{eq:search-width}
\end{equation}

\noindent\textbf{Evidence-Guided $(C,W,L)$ Joint Synthesis.}
The parent's bottlenecks show where to revise, and their witnesses show what went wrong there: a missing observation implicates $C$, a misordered routine $W$, late feedback $L$ (\hyperref[sec:problem-formulation]{Sec.~\ref*{sec:problem-formulation}}). Each of $K_t$ outer-agent instances $\mathcal A_{\pi_{\mathrm{FM}}}$ reads the parent, topology, and witnesses $\mathcal Z_t(\mathcal U_t)$ to propose an edit:
\nopagebreak
\begin{equation}
    \Delta H_{t,k}=\mathcal A_{\pi_{\mathrm{FM}}}\!\left(H_t^{\mathrm{parent}};\Phi_t,\mathcal Z_t(\mathcal U_t)\right).
\label{eq:evidence-guided-synthesis}
\end{equation}
Applying the edit gives $H_{t,k}=H_t^{\mathrm{parent}}\oplus\Delta H_{t,k}$. Final selection follows Eq.~\ref{eq:parent-selection}; successful ties prefer the smallest $|\Delta H|$ for compact distillation.

\subsubsection{Topology-Guided Meta-Distillation}
\label{sec:shared-distillation}

\methodinsight{3}{\textcolor{echoaccent}{\textbf{What is worth transferring?}} Past experience transfers as modular causal units sharing physical preconditions with the student. These verified mechanisms constitute the ``meta'' of meta-distillation, transplanted right where target execution stalls.}

\noindent\textbf{Modular Causal Units.}
Local transitions such as pre-grasp alignment, surface stabilization, and unlatching recur across tasks with different topologies. An edge defines a \emph{causal unit} $u=(\mathcal I_u,\mathcal O_u,\mathcal P_u)\in\mathcal U$ with newly established effects
\nopagebreak
\begin{equation}
    E_u=\{a\in\mathcal O_u:\mathcal I_u\not\models a\},
\label{eq:causal-unit}
\end{equation}
where $\mathcal I_u$ and $\mathcal O_u$ are the conditions before and after the transition, $\mathcal P_u$ is the code that realizes it, and $E_u$ collects its newly established outcomes. A condition $a=(p,\boldsymbol x,V)$ requires a property $p$ of objects $\boldsymbol x$, such as position, to lie in $V$, and a state $S$ supports it ($S\models a$) when the observed property does. Units in the experience bank $\mathcal B$ supported by achieved transition witnesses form $\mathcal U^+$. Units of different specialists that establish matching conditions under an object mapping can merge into one implementation.

\noindent\textbf{Frontier Dependency-Gap Matching.}
Failures of the student $H_t$ on query tasks $\mathcal T_{\mathrm q}$ of the task set $\mathcal T$ show what to retrieve. Each failed trial $c$ has a frontier $F_t(c)$ on its task topology $\Phi_i$ (\hyperref[eq:reprojection-frontier]{Eq.~(\ref*{eq:reprojection-frontier})}). The dependency gap of a transition $q$ entering $F_t(c)$ is the part of its outcomes $E_q$ unsupported by the state $S_t(c)$ observed in the failed run:
\nopagebreak
\begin{equation}
    D_t(q,c)=\{a\in E_q \mid S_t(c)\not\models a\}.
\label{eq:dependency-gap}
\end{equation}
A bank unit $u \in \mathcal U^+$, mapped to the student's objects by $\sigma$, covers the part of the gap that its outcomes establish:
\nopagebreak
\begin{equation}
\begin{aligned}
    C_t(q,u,\sigma;c)&=\bigl\{d\in D_t(q,c) \mid \sigma(E_u|_{\operatorname{dom}\sigma})\models d\bigr\},\\
    R_t(q,c)&=\bigl\{(u,\sigma):C_t(q,u,\sigma;c)\ne\varnothing\bigr\}.
\end{aligned}
\label{eq:topology-alignment}
\end{equation}
Host code retrieves candidates by outcome coverage. The outer agent adapts code to target entry conditions and validates effects in rollouts (Supp.~\ref{sec:supp-method-implementation}).

\noindent\textbf{Shared Meta-Distillation and Continual Transfer.}
On trial panel $\Xi_{\mathcal T}$, the development loss scores shared harness $H$ by success, retained progress, and edit size:
\nopagebreak
\begin{equation}
\begin{aligned}
    \mathcal L_{\mathrm{meta}}(H) &= \bigl(1-\bar S(H)\bigr)
    +\lambda_T\bigl(1-\bar p(H)\bigr) \\
    &\quad +\lambda_S\bigl[\,|\Delta H|-b_{\mathcal B}\,\bigr]_+,
\end{aligned}
\label{eq:shared-distillation}
\end{equation}
where $\bar S$ averages official success and $\bar p$ measures retained progress on fixed source reference topologies over $\mathcal T$ (\hyperref[eq:retained-progress]{Eq.~(\ref*{eq:retained-progress})}). Here $|\Delta H|$ is the $(C,W,L)$ code size relative to $H_0$. Penalizing code beyond $b_{\mathcal B}$, the mean reference-harness increment, favors compact shared code over simply concatenating specialists.

The outer agent acts as a black-box optimizer, proposing the next harness from the loss, evaluations, $\mathcal Q_t$, and bank:
\nopagebreak
\begin{equation}
    H_{t+1} = \mathcal A_{\pi_{\mathrm{FM}}}\!\bigl(H_t;\ \mathcal L_t,\ \mathcal E_t,\ \mathcal Q_t,\ \mathcal B\bigr).
\label{eq:harness-update}
\end{equation}
Optimization stops when loss drops below $\tau$ or after $R_{\max}$ rounds, returning best $H_{\mathrm{sh}}^*$.

The loop serves two uses. Merging sets $H_{\mathrm{init}}{=}H_0$ and $\mathcal T_{\mathrm q}{=}\mathcal T$ with the specialists in $\mathcal B$, and freezes the result for unseen tasks. Continual learning starts from the previous shared harness with new tasks as $\mathcal T_{\mathrm q}$, whose topologies query the bank before joining it.

\begin{figure}[t]
  \centering
  \includegraphics[width=\columnwidth]{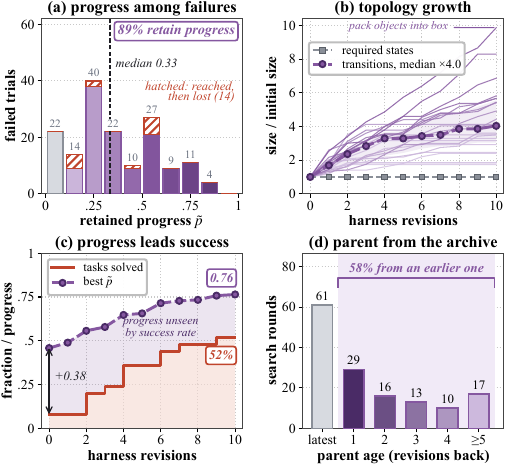}
  \caption{\textbf{Topology dynamics} across 25 RoboDojo tasks ($\leq$10 revisions). \textbf{\textit{(a)}} Terminal failure progress. \textbf{\textit{(b)}} Relative graph size. \textbf{\textit{(c)}} Solved ratio and best progress. \textbf{\textit{(d)}} Parent age distribution.}
  \label{fig:topology-read}
\end{figure}

\setcounter{dbltopnumber}{1}
\begin{table*}[t]
\begingroup
\input{tables/main_results}
\endgroup
\par\medskip
\input{tables/robodojo_results}
\end{table*}

\begin{table*}[t]
\input{tables/ablation_distillation}
\par\medskip
\begingroup
\captionsetup{type=figure}
\input{figures/src/libero_case_study/figure}
\endgroup
\end{table*}

\begin{table}[t]
\input{tables/ablation_main}
\end{table}

\begin{table}[t]
\input{tables/model_choices}
\par\medskip
\input{tables/simulator_alignment}
\end{table}

\subsection{Novelty Statement}
\label{sec:method-novelty}
\noindent\textbf{Emergent topology as explicit meta-knowledge.}
Prior systems revise harnesses and skills~\cite{wang2026self,wang2026harnesspai,wang2026adarsi,yu2026robodoctor}, or evolve stateful code~\cite{hu2026embodied,gao2026selfevolving}. \methodname{} uses the emergent topology as explicit meta-knowledge, separating invariant physical traces from revisable prerequisites. Incoming evidence can retrospectively re-score archived trials, uncovering latent progress and localized regressions for parent selection without collecting new physical rollouts.

\noindent\textbf{Loss-grounded unified meta-distillation.}
Prior methods distill playbooks~\cite{kim2026recursive}, reuse skills and code~\cite{lu2026aspire,li2026embodiedevo,wang2026harnesspai,hu2026embodied}, or route harnesses~\cite{liu2026adaptive}, but do not jointly optimize task success, progress retention, and code parsimony. \methodname{} makes this trade-off explicit in its meta-distillation loss. One optimization loop unifies multi-specialist merging and continual transfer, retrieving verified causal units by preconditions and effects to bridge topological frontier gaps.
\endgroup

%% file: tables/main_results.tex
\centering
\caption{LIBERO-PRO SR (\%)~\cite{zhou2025libero}. \methodname{} gains $9.75$ SR points over $H_0$, reaching $98.75\%$. Optimization yields the largest gains on Long tasks. Details and inference costs: Supp.~\ref{sec:supp-additional-results}.}
\label{tab:main-results}
\footnotesize
\setlength{\tabcolsep}{2.2pt}
\setlength{\aboverulesep}{0pt}
\setlength{\belowrulesep}{0pt}
\renewcommand{\arraystretch}{1.0}
\begin{tabularx}{\textwidth}{@{}l c c c c *{4}{>{\centering\arraybackslash}X} c c@{}}
\toprule
\multicolumn{1}{c}{\multirow{2.2}{*}{\textbf{Method}}} & \multirow{2.2}{*}{\textbf{Year}} & \multirow{2.2}{*}{\textbf{Model (effort)}} & \multirow{2.2}{*}{\textbf{Train}} & \multirow{2.2}{*}{\textbf{Optimize}} & \multicolumn{4}{c}{\textbf{Success rate (\%) $\uparrow$, position / task}} & \multicolumn{2}{c}{\textbf{Average $\uparrow$}} \\
\cmidrule(lr){6-9} \cmidrule(lr){10-11}
 & & & & & \textbf{Spatial} & \textbf{Object} & \textbf{Goal} & \textbf{Long} & \textbf{6 suites} & \textbf{8 suites} \\
\midrule
\rowcolor{tablegray}\multicolumn{11}{l}{\rule{0pt}{2.0ex}\textbf{E2E models}}\\
$\pi_{0.5}$~\cite{intelligence2025pi_} & 2025 & - & \yes & \no & \scorepair{20.0}{1.0} & \scorepair{17.0}{1.0} & \scorepair{38.0}{2.0} & \scorepair{8.0}{1.0} & 13.2 & 11.0 \\
$\pi_{\mathrm{RLinf}}$~\cite{zang2025rlinf} & 2025 & - & \yes & \no & \scorepair{59.0}{42.0} & \scorepair{78.0}{71.0} & \scorepair{42.0}{45.0} & \scorepair{14.0}{49.0} & 56.2 & 50.0 \\
AtomVLA~\cite{sun2026atomvla} & 2026 & - & \yes & \no & \scorepair{16.0}{1.0} & \scorepair{10.0}{0.0} & \scorepair{2.0}{11.0} & \scorepair{1.0}{9.0} & 6.7 & 6.3 \\
X-VLA~\cite{zheng2026x} & 2025 & - & \yes & \no & \scorepair{0.0}{0.0} & \scorepair{2.0}{8.0} & \scorepair{1.0}{9.0} & \scorepair{0.0}{10.0} & 3.3 & 3.8 \\
\midrule
\rowcolor{tablegray}\multicolumn{11}{l}{\rule{0pt}{2.0ex}\textbf{Agents}}\\
OpenRUA~\cite{chu2026openrua} & 2026 & Opus 5 (high) & \no & \no & \scorepair{98.0}{\bestval{100.0}} & \scorepair{98.0}{97.0} & \scorepair{79.0}{79.0} & \scorepair{80.0}{65.0} & 91.8 & 87.0 \\
RPent~\cite{zhang2026harness} & 2026 & GPT-6 Astra (low) & \no & \yes & \scorepair{98.0}{\bestval{100.0}} & \scorepair{\runnerupval{99.0}}{\bestval{100.0}} & \scorepair{\runnerupval{99.0}}{88.0} & \scorepair{72.0}{85.0} & 97.33 & 92.63 \\
ASPIRE~\cite{lu2026aspire} & 2026 & GPT-6 Astra (xhigh) & \no & \yes & \scorepair{\runnerupval{98.6}}{\runnerupval{97.6}} & \scorepair{\runnerupval{99.0}}{\runnerupval{99.4}} & \scorepair{97.0}{\runnerupval{96.8}} & \scorepair{\runnerupval{84.6}}{\runnerupval{95.0}} & \runnerupval{98.07} & \runnerupval{96.00} \\
\rowcolor{tableaccent}
\methodlabel{\methodname{} $H_0$} & 2026 & GPT-6 Astra (low) & \no & \no & \scorepair{93.0}{\bestval{100.0}} & \scorepair{\bestval{100.0}}{99.0} & \scorepair{97.0}{90.0} & \scorepair{72.0}{61.0} & 96.50 & 89.00 \\
\rowcolor{tableaccent}
\methodlabel{\methodname{} (Ours)} & 2026 & GPT-6 Astra (low) & \no & \yes & \scorepair{\bestval{99.0}}{\bestval{100.0}} & \scorepair{\bestval{100.0}}{\bestval{100.0}} & \scorepair{\bestval{100.0}}{\bestval{98.0}} & \scorepair{\bestval{96.0}}{\bestval{97.0}} & \bestval{99.50} & \bestval{98.75} \\
\rowcolor{deltaselected!35}
\multicolumn{5}{c}{\textbf{$\Delta$ vs.\ $H_0$}} & \mbox{\deltagain{6.0} / \textbf{0.0}} & \mbox{\textbf{0.0} / \deltagain{1.0}} & \mbox{\deltagain{3.0} / \deltagain{8.0}} & \mbox{\deltagain{24.0} / \deltagain{36.0}} & \deltagain{3.00} & \deltagain{9.75} \\
\bottomrule
\end{tabularx}

%% file: tables/robodojo_results.tex
\centering
\caption{RoboDojo Score/SR (\%)~\cite{chen2026robodojo}. \methodname{} gains $30.42$/$32.75$ points over $H_0$. Harness optimization improves both metrics across all capability groups. Details and inference costs: Supp.~\ref{sec:supp-additional-results}.}
\label{tab:robodojo-results}
\footnotesize
\setlength{\tabcolsep}{1.3pt}\setlength{\arraycolsep}{1.3pt}
\setlength{\aboverulesep}{0pt}
\setlength{\belowrulesep}{0pt}
\renewcommand{\arraystretch}{1.0}
\begin{tabularx}{\textwidth}{@{}l c c c *{7}{>{\centering\arraybackslash}X}@{}}
\toprule
\multicolumn{1}{c}{\multirow{2.2}{*}{\textbf{Method}}} & \multirow{2.2}{*}{\textbf{Year}} & \multirow{2.2}{*}{\textbf{Train}} & \multirow{2.2}{*}{\textbf{Optimize}} & \multicolumn{2}{c}{\textbf{Generalization $\uparrow$}} & \multicolumn{4}{c}{\textbf{Capability $\uparrow$}} & \multirow{2}{*}{\textbf{Average $\uparrow$}} \\
\cmidrule(lr){5-6}\cmidrule(lr){7-10}
 & & & & \textbf{Std.} & \textbf{Random} & \textbf{Precision} & \textbf{Long} & \textbf{Memory} & \textbf{Open} &  \\
\midrule
\rowcolor{tablegray}\multicolumn{11}{l}{\rule{0pt}{2.0ex}\textbf{E2E models}}\\
$\pi_{0.5}$~\cite{intelligence2025pi_} & 2025 & \yes & \no & \scorepair{20.93}{14.89} & \scorepair{5.82}{1.44} & \scorepair{12.40}{5.50} & \scorepair{23.54}{14.67} & \scorepair{5.89}{4.67} & \scorepair{1.98}{1.67} & \scorepair{11.44}{6.93} \\
Awomo-0.5~\cite{team2026awomo} & 2026 & \yes & \no & \scorepair{37.70}{30.89} & \scorepair{22.92}{16.78} & \scorepair{39.65}{33.50} & \scorepair{40.36}{26.75} & \scorepair{42.27}{41.22} & \scorepair{24.09}{22.92} & \scorepair{35.34}{29.64} \\
VPP2~\cite{guo2026video} & 2026 & \yes & \no & \scorepair{\bestval{52.14}}{\bestval{46.44}} & \scorepair{20.63}{12.56} & \scorepair{\runnerupval{43.03}}{\runnerupval{35.17}} & \scorepair{\bestval{56.28}}{37.00} & \scorepair{56.01}{55.56} & \scorepair{4.61}{4.08} & \scorepair{39.26}{32.26} \\
\rowcolor{tablegray}\multicolumn{11}{l}{\rule{0pt}{2.0ex}\textbf{Self-optimizing agents}}\\
PhysicalRSI~\cite{physicalrsi2026} & 2026 & \no & \yes & \scorepair{31.57}{25.67} & \scorepair{11.04}{5.56} & \scorepair{38.05}{32.50} & \scorepair{46.37}{\runnerupval{37.33}} & \scorepair{46.74}{46.56} & \scorepair{28.88}{24.92} & \scorepair{36.27}{31.38} \\
EmbodiedEvo~\cite{li2026embodiedevo} & 2026 & \no & \yes & \scorepair{34.55}{30.33} & \scorepair{35.88}{\runnerupval{31.33}} & \scorepair{25.15}{16.25} & \scorepair{33.83}{21.75} & \scorepair{79.40}{76.67} & \scorepair{51.81}{49.25} & \scorepair{45.08}{38.95} \\
WorldScape-R$^2$SI~\cite{team2026r} & 2026 & \no & \yes & \scorepair{\runnerupval{46.62}}{40.33} & \scorepair{\runnerupval{39.28}}{\runnerupval{31.33}} & \scorepair{23.51}{15.50} & \scorepair{51.48}{34.25} & \scorepair{65.78}{63.67} & \scorepair{49.16}{47.50} & \scorepair{46.58}{39.35} \\
HexaShell-Preview~\cite{gao2026hexashell} & 2026 & \no & \yes & \scorepair{43.28}{38.00} & \scorepair{35.15}{31.00} & \scorepair{41.25}{32.00} & \scorepair{42.91}{31.50} & \scorepair{\runnerupval{84.53}}{\runnerupval{81.00}} & \scorepair{\runnerupval{57.97}}{\runnerupval{55.25}} & \scorepair{\runnerupval{53.18}}{\runnerupval{46.85}} \\
\midrule
\rowcolor{tableaccent}
\methodlabel{\methodname{} $H_0$} & 2026 & \no & \no & \scorepair{26.08}{21.67} & \scorepair{20.25}{15.00} & \scorepair{19.94}{15.00} & \scorepair{27.81}{6.25} & \scorepair{60.00}{56.67} & \scorepair{34.13}{31.25} & \scorepair{33.01}{25.50} \\
\rowcolor{tableaccent}
\methodlabel{\methodname{} (Ours)} & 2026 & \no & \yes & \scorepair{45.50}{\runnerupval{41.67}} & \scorepair{\bestval{47.08}}{\bestval{41.67}} & \scorepair{\bestval{48.56}}{\bestval{42.50}} & \scorepair{\runnerupval{54.63}}{\bestval{42.50}} & \scorepair{\bestval{93.33}}{\bestval{93.33}} & \scorepair{\bestval{74.31}}{\bestval{71.25}} & \scorepair{\bestval{63.43}}{\bestval{58.25}} \\
\rowcolor{deltaselected!35}
\multicolumn{4}{c}{\textbf{$\Delta$ vs.\ $H_0$}} & \deltapair{19.42}{20.00} & \deltapair{26.83}{26.67} & \deltapair{28.63}{27.50} & \deltapair{26.81}{36.25} & \deltapair{33.33}{36.67} & \deltapair{40.19}{40.00} & \deltapair{30.42}{32.75} \\
\bottomrule
\end{tabularx}

%% file: tables/ablation_distillation.tex
\centering
\caption{\textbf{Shared distillation (A) and continual learning (B)} on LIBERO-PRO ($8$ source/new tasks $\times 10$ test seeds). \methodname{} gains $10.0$ unseen-task SR points over direct distillation. Yellow: selected rounds or replay minima; purple: target SR; superscripts: gains.}
\label{tab:ablation-distillation}
\label{tab:continual-learning}
\footnotesize
\setlength{\tabcolsep}{0.5pt}
\setlength{\aboverulesep}{0pt}
\setlength{\belowrulesep}{0pt}
\renewcommand{\arraystretch}{1.13}
\newcommand{\setting}[2]{\textsc{#1}$_{#2}$}
\definecolor{distillselected}{HTML}{F5C75C}
\newcommand{\srheat}[2]{\cellcolor{echoaccent!#1}#2}
\newcommand{\selected}[1]{\cellcolor{distillselected!35}\textbf{#1}}
\newcommand{\srgain}[2]{\mbox{\bestval{#1}\textsuperscript{\fontsize{4.8}{5}\selectfont\textcolor{ablationgain}{+#2}}}}
\begin{tabularx}{\textwidth}{@{} >{\raggedright\arraybackslash}p{0.125\textwidth} *{16}{>{\centering\arraybackslash}X} @{}}
\toprule
\tabgroup{17}{A. Source scaling and shared distillation}
\multicolumn{1}{c}{\multirow{2}{*}{\textbf{Setting}}} & \multicolumn{5}{c}{\textbf{Development loss $\downarrow$}} & \multicolumn{3}{c}{\textbf{Selected $H$}} & \multicolumn{5}{c}{\textcolor{echoaccent}{\textbf{Target test SR (\%) $\uparrow$}}} & \multicolumn{2}{c}{\textbf{Dev. effort $\downarrow$}} & \multirow{2}{*}{$\frac{\Delta H}{\overline{\Delta H^{*}}}\!\downarrow$} \\
\cmidrule(lr){2-6}\cmidrule(lr){7-9}\cmidrule(lr){10-14}\cmidrule(lr){15-16}
 & R1 & R2 & R3 & R4 & R5 & SR$\uparrow$ & Prog.$\uparrow$ & Loss$\downarrow$ & Spatial & Object & Goal & Long & Avg. & Eps. & Tokens & \\
\midrule
$H_0$ & - & - & - & - & - & - & 69.8 & 0.802 & \srheat{0}{0.0} & \srheat{18}{35.0} & \srheat{0}{0.0} & \srheat{0}{0.0} & \srheat{4}{8.8} & 24 & 0.00 & 0.00 \\
\textcolor{echoaccent}{\setting{Ours}{2}} & 0.671 & 0.288 & \selected{0.014} & - & - & 95.0 & 100.0 & 0.014 & \srheat{48}{95.0} & \srheat{22}{45.0} & \srheat{8}{15.0} & \srheat{2}{5.0} & \srheat{20}{40.0} & 24 & 17.85 & 1.38 \\
\textcolor{echoaccent}{\setting{Ours}{4}} & 0.434 & 0.160 & \selected{0.135} & 0.385 & 0.250 & 80.0 & 97.2 & 0.135 & \srheat{35}{70.0} & \srheat{35}{70.0} & \srheat{0}{0.0} & \srheat{8}{15.0} & \srheat{19}{38.8} & 72 & 27.69 & 1.84 \\
\textcolor{echoaccent}{\setting{Ours}{6}} & 0.703 & 0.272 & \selected{0.173} & 0.185 & 0.208 & 95.0 & 96.3 & 0.173 & \srheat{45}{90.0} & \srheat{50}{100.0} & \srheat{2}{5.0} & \srheat{0}{0.0} & \srheat{24}{48.8} & 108 & 31.89 & 2.04 \\
\setting{w/o topology}{8} & 0.618 & 0.296 & 0.601 & 0.151 & \selected{0.151} & 88.8 & 95.8 & 0.151 & \srheat{30}{60.0} & \srheat{48}{95.0} & \srheat{0}{0.0} & \srheat{5}{10.0} & \srheat{21}{41.3} & 144 & 47.72 & 2.05 \\
\rowcolor{tableaccent}
\methodlabel{\setting{Ours}{8}} & 0.652 & 0.241 & 0.554 & 0.293 & \selected{0.237} & 86.3 & 93.1 & 0.237 & \srheat{48}{95.0} & \srheat{48}{95.0} & 10.0 & 5.0 & \srheat{26}{\srgain{51.3}{10.0}} & 144 & 41.72 & 2.75 \\
\midrule
\tabgroup{17}{B. Continual learning}
\textcolor{echoaccent}{\setting{Ours}{2}} & - & - & - & - & - & 100.0 & 100.0 & 0.000 & \srheat{48}{95.0} & \srheat{48}{95.0} & \srheat{5}{10.0} & \srheat{2}{5.0} & \srheat{26}{51.3} & 6 & 5.10 & 2.75 \\
\textcolor{echoaccent}{\setting{Ours}{4}} & - & - & - & - & - & 97.2 & 99.1 & 0.028 & \srheat{48}{95.0} & \srheat{48}{95.0} & \srheat{5}{10.0} & \srheat{2}{5.0} & \srheat{26}{51.3} & 6 & 5.71 & 2.75 \\
\textcolor{echoaccent}{\setting{Ours}{6}} & \selected{0.124} & 0.176 & 0.224 & 0.177 & 0.132 & 88.1 & 90.5 & 0.124 & \srheat{48}{95.0} & \srheat{50}{100.0} & \srheat{5}{10.0} & \srheat{10}{20.0} & \srheat{28}{56.3} & 36 & 53.93 & 2.94 \\
\setting{From Scratch}{8} & 0.757 & 0.514 & 0.351 & \selected{0.234} & 0.234 & 87.5 & 88.9 & 0.148 & \srheat{28}{55.0} & \srheat{42}{85.0} & \srheat{8}{15.0} & \srheat{30}{60.0} & \srheat{27}{53.8} & 26.25 & 32.96 & 3.71 \\
\rowcolor{tableaccent}
\methodlabel{\setting{Ours}{8}} & 0.378 & 0.193 & 0.174 & \selected{0.076} & - & 93.8 & 93.8 & 0.076 & \srheat{45}{90.0} & \srheat{45}{90.0} & 20.0 & \srheat{50}{100.0} & \srheat{38}{\srgain{75.0}{21.2}} & 30 & 41.46 & 3.50 \\
\bottomrule
\end{tabularx}

%% file: figures/src/libero_case_study/figure.tex
  \centering
  \includegraphics[width=\textwidth,trim=10bp 0 1bp 0,clip]{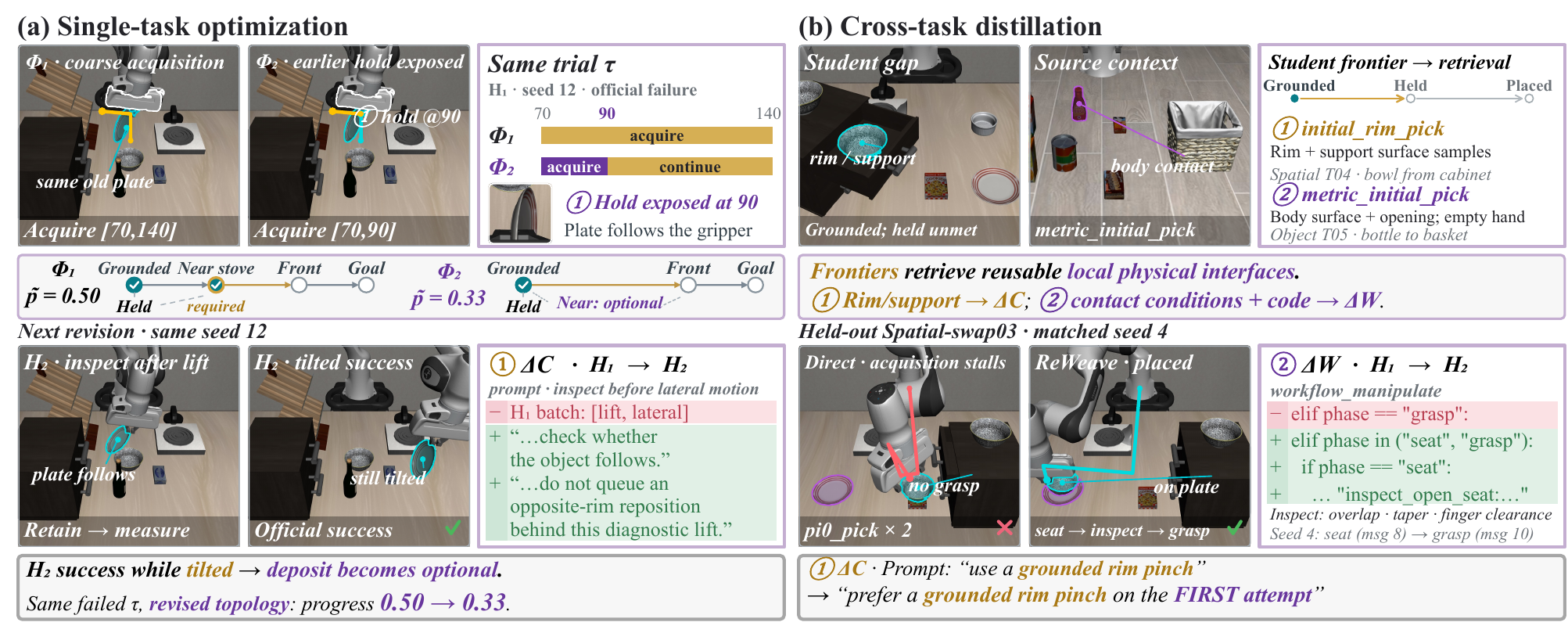}
  \caption{\textbf{Re-reading and transferring experience on LIBERO-PRO.} \textbf{\textit{(a)}} An archived trial reprojected under revised evidence, exposing earlier grasp retention. \textbf{\textit{(b)}} Retrieved units and resulting harness revisions in a shared lineage evaluated on held-out seed~4.}
  \label{fig:libero-case-study}

%% file: tables/ablation_main.tex
\centering
\caption{Harness ablations. Avg./Dev.: SR (\%). $\Delta$ vs.\ Full.}
\label{tab:ablation-main}
\newcommand{\metricdelta}[3]{\mbox{#1\textsuperscript{\fontsize{5.3}{6}\selectfont\textcolor{#3}{#2}}}}
\footnotesize
\setlength{\tabcolsep}{0.2pt}
\setlength{\aboverulesep}{0pt}
\setlength{\belowrulesep}{0pt}
\renewcommand{\arraystretch}{1.04}
\begin{tabularx}{\linewidth}{@{}>{\raggedright\arraybackslash}X
*{6}{c}@{}}
\toprule
\multicolumn{1}{c}{\multirow{2}{*}{\textbf{Variant}}} & \multicolumn{1}{c}{\multirow{2}{*}{\textbf{Avg.$\uparrow$}}} & \multicolumn{1}{c}{\multirow{2}{*}{\textbf{Dev.$\uparrow$}}} & \multicolumn{2}{c}{\textbf{Search cost $\downarrow$}} & \multicolumn{2}{c}{\textbf{Rollout cost $\downarrow$}} \\
\cmidrule(lr){4-5} \cmidrule(lr){6-7}
 & & & \multicolumn{1}{c}{Rounds} & \multicolumn{1}{c}{Trials} & \multicolumn{1}{c}{Dec.} & \multicolumn{1}{c}{Steps} \\
\midrule
\tabgroup{7}{A. Solution space}
$H_0$ & \metricdelta{3.75}{-51.25}{ablationdrop} & \metricdelta{4.17}{-70.83}{ablationdrop} & - & - & \metricdelta{6.3}{-5.7}{ablationgain} & \metricdelta{313.1}{+38.9}{ablationdrop} \\
$C$ & \metricdelta{47.50}{-7.50}{ablationdrop} & \metricdelta{66.67}{-8.33}{ablationdrop} & \metricdelta{3.6}{-0.2}{ablationgain} & \metricdelta{10.9}{-0.5}{ablationgain} & \metricdelta{11.2}{-0.8}{ablationgain} & \metricdelta{277.9}{+3.7}{ablationdrop} \\
$C + W$ & \metricdelta{52.50}{-2.50}{ablationdrop} & \metricdelta{79.17}{+4.17}{ablationgain} & \metricdelta{3.5}{-0.3}{ablationgain} & \metricdelta{10.8}{-0.6}{ablationgain} & \metricdelta{12.3}{+0.3}{ablationdrop} & \metricdelta{277.4}{+3.2}{ablationdrop} \\
\rowcolor{tableaccent}
\methodlabel{$\boldsymbol{C + W + L}$} & \metricdelta{55.00}{0.00}{black!50} & \metricdelta{75.00}{0.00}{black!50} & \metricdelta{3.8}{0.0}{black!50} & \metricdelta{11.4}{0.0}{black!50} & \metricdelta{12.0}{0.0}{black!50} & \metricdelta{274.2}{0.0}{black!50} \\
\midrule
\tabgroup{7}{B. Harness optimization strategies}
w/o topology & \metricdelta{46.25}{-8.75}{ablationdrop} & \metricdelta{62.50}{-12.50}{ablationdrop} & \metricdelta{3.9}{+0.1}{ablationdrop} & \metricdelta{11.6}{+0.2}{ablationdrop} & \metricdelta{12.1}{+0.1}{ablationdrop} & \metricdelta{274.4}{+0.2}{ablationdrop} \\
w/ fixed topology & \metricdelta{41.25}{-13.75}{ablationdrop} & \metricdelta{75.00}{0.00}{black!50} & \metricdelta{3.6}{-0.2}{ablationgain} & \metricdelta{11.3}{-0.1}{ablationgain} & \metricdelta{12.7}{+0.7}{ablationdrop} & \metricdelta{287.2}{+13.0}{ablationdrop} \\
w/o~reprojection & \metricdelta{47.50}{-7.50}{ablationdrop} & \metricdelta{62.50}{-12.50}{ablationdrop} & \metricdelta{4.4}{+0.6}{ablationdrop} & \metricdelta{13.6}{+2.2}{ablationdrop} & \metricdelta{15.0}{+3.0}{ablationdrop} & \metricdelta{283.9}{+9.7}{ablationdrop} \\
\rowcolor{tableaccent}
\methodlabel{\methodname{}} & \metricdelta{55.00}{0.00}{black!50} & \metricdelta{75.00}{0.00}{black!50} & \metricdelta{3.8}{0.0}{black!50} & \metricdelta{11.4}{0.0}{black!50} & \metricdelta{12.0}{0.0}{black!50} & \metricdelta{274.2}{0.0}{black!50} \\
\midrule
\tabgroup{7}{C. Search configuration}
$K{=}1$ & \metricdelta{55.00}{0.00}{black!50} & \metricdelta{75.00}{0.00}{black!50} & \metricdelta{3.8}{0.0}{black!50} & \metricdelta{11.4}{0.0}{black!50} & \metricdelta{12.0}{0.0}{black!50} & \metricdelta{274.2}{0.0}{black!50} \\
Fixed $K{=}2$ & \metricdelta{52.50}{-2.50}{ablationdrop} & \metricdelta{91.67}{+16.67}{ablationgain} & \metricdelta{3.0}{-0.8}{ablationgain} & \metricdelta{18.1}{+6.7}{ablationdrop} & \metricdelta{12.0}{0.0}{black!50} & \metricdelta{284.4}{+10.2}{ablationdrop} \\
Fixed $K{=}3$ & \metricdelta{56.25}{+1.25}{ablationgain} & \metricdelta{83.33}{+8.33}{ablationgain} & \metricdelta{3.1}{-0.7}{ablationgain} & \metricdelta{28.1}{+16.7}{ablationdrop} & \metricdelta{11.7}{-0.3}{ablationgain} & \metricdelta{273.2}{-1.0}{ablationgain} \\
\rowcolor{tableaccent}
\methodlabel{Dynamic $\kappa{=}30$} & \metricdelta{55.00}{0.00}{black!50} & \metricdelta{83.33}{+8.33}{ablationgain} & \metricdelta{2.9}{-0.9}{ablationgain} & \metricdelta{16.9}{+5.5}{ablationdrop} & \metricdelta{12.0}{0.0}{black!50} & \metricdelta{283.0}{+8.8}{ablationdrop} \\
\bottomrule
\end{tabularx}

%% file: tables/model_choices.tex
\centering
\caption{Model ablations for held-out SR (\%).}
\label{tab:model-choices}
\footnotesize
\setlength{\tabcolsep}{3pt}
\renewcommand{\arraystretch}{1.03}
\begin{tabularx}{\linewidth}{@{}>{\centering\arraybackslash}X >{\centering\arraybackslash}X c@{}}
\toprule
\textbf{Online LM} & \textbf{Offline LM} & \textbf{SR (\%) $\uparrow$} \\
\midrule
\tabgroup{3}{A. Online execution with frozen harnesses}
\mbox{GPT-6.1 Terra (high)} & \mbox{GPT-6.1 Sol (xhigh)} & 55.00 \\
\mbox{Claude Sonnet 5.5 (high)} & \mbox{GPT-6.1 Sol (xhigh)} & 57.50 \\
\mbox{DeepSeek V4.1 Flash (high)} & \mbox{GPT-6.1 Sol (xhigh)} & 30.00 \\
\midrule
\tabgroup{3}{B. Offline optimization with Terra execution}
\mbox{GPT-6.1 Terra (high)} & \mbox{GPT-6.1 Sol (xhigh)} & 55.00 \\
\mbox{GPT-6.1 Terra (high)} & \mbox{Claude Sonnet 5.5 (xhigh)} & 92.50 \\
\mbox{GPT-6.1 Terra (high)} & \mbox{DeepSeek V4.1 Flash (xhigh)} & 65.00 \\
\bottomrule
\end{tabularx}

%% file: tables/simulator_alignment.tex
\centering
\caption{\textbf{Scoring fidelity and stability} on LIBERO-PRO.}
\label{tab:simulator-alignment}
\footnotesize
\setlength{\tabcolsep}{2pt}
\setlength{\aboverulesep}{0pt}
\setlength{\belowrulesep}{0pt}
\renewcommand{\arraystretch}{1.08}
\begin{tabularx}{\linewidth}{@{}>{\raggedright\arraybackslash}X >{\centering\arraybackslash}p{62pt} >{\centering\arraybackslash}p{80pt}@{}}
\toprule
\textbf{Evaluation metric} & \textbf{Agent scorer} & \textbf{Sim reference} \\
\midrule
\tabgroup{3}{A. Milestone state grounding (39 failure traces)}
Milestone precision / recall & 100.0\% / 100.0\% & 48 / 48 resolved states \\
Progress MAE on failures & 0.000 & Ground truth ($n{=}30$) \\
\midrule
\tabgroup{3}{B. Optimization guidance \& stability ($H_0 \to H_5$)}
Search progress gain & +29.17 pp & [+25.00, +32.64] pp \\
Regret (physical / full) & 0.000 / 0.000 & $\le 0.025$ / $\le 0.053$ \\
5-pass repeat stability & 5 / 5 select $H_5$ & Score $\text{SD} = 0.000$ \\
\bottomrule
\end{tabularx}

%% file: sections/05_experiments.tex
\begingroup
\setcounter{secnumdepth}{2}
\clubpenalty=10000
\renewcommand{\finding}[2]{\textbf{\textit{(#1)~#2}}\enspace}
\section{Experiments}
\label{sec:experiments}

\subsection{Experimental Setup}
\label{sec:exp-setup}
\noindent{\textbf{(1) Evaluation details.}} We evaluate LIBERO-PRO~\cite{zhou2025libero} (80 tasks, four suites) and RoboDojo~\cite{chen2026robodojo} (42 tasks, five suites), following development/test separation~\cite{lu2026aspire,zhang2026harness,ding2026zetta} (seed~0 for development, at most 10 rounds per task; seeds~1-10 for held-out test). Policies and SAM3 are frozen. Tables~\ref{tab:main-results}-\ref{tab:robodojo-results} report system-level performance and gains over $H_0$. For ablation and transfer, Tables~\ref{tab:ablation-distillation}-\ref{tab:model-choices} use eight-task MiniSets of each category's hardest tasks at $1\times$ environment-step budgets. We report SR, retained progress, trials, and costs (\mbox{Supp.~\ref{sec:supp-experimental-setup}}). See Supp.~\ref{sec:supp-additional-results} for paired comparisons, confidence intervals, and an independent optimization rerun.

\noindent{\textbf{(2) Fairness and comparison protocol.}} Existing embodied agents employ disparate backbones, undisclosed trial budgets, and bespoke prompt tuning~\cite{lu2026aspire,zhang2026harness,gao2026hexashell}; re-evaluating all proprietary systems is cost-prohibitive. We adopt an analytical protocol: assessing optimization and transfer through controlled $H_0$ baselines and structural ablations. Supp.~\ref{sec:supp-experimental-setup} details fairness, compute accounting, and comparison scope.

\subsection{Benchmark Results}
\label{sec:main-results}
\label{sec:robodojo-results}
\textbf{\textit{(1)~Optimization gains.}} On LIBERO-PRO and RoboDojo, \methodname{} reaches $98.75\%$ and $58.25\%$ SR, respectively, up $9.75$ and $32.75$ points over $H_0$ (Tables~\ref{tab:main-results}-\ref{tab:robodojo-results}). Long gains $24$/$36$ points on LIBERO-PRO; RoboDojo improves in every group. \finding{2}{A capable seed harness.} $H_0$ reaches $89.00\%$/$25.50\%$ SR ($96.50\%$ six-suite LIBERO-PRO average) with the same execution models and embodiment interfaces; optimization changes the surrounding harness.

\subsection{Cross-Task Transfer}
\label{sec:ablation-distillation}
\label{sec:continual-learning}
\noindent{\textbf{Unseen task generalization.}}
\finding{1}{Source fit does not guarantee transfer.} With eight specialists and matched seed, optimizer, loss, and development data, topology-guided retrieval reaches $51.3\%$ unseen SR versus $41.3\%$ without topology, despite lower source SR ($86.3\%$ vs.\ $88.8\%$; Table~\ref{tab:ablation-distillation}A). Gains are concentrated in Spatial ($95\%$ vs.\ $60\%$), using fewer offline tokens. SR with two, four, six, and eight sources is $40.0\%$, $38.8\%$, $48.8\%$, and $51.3\%$, respectively, showing nonmonotonic transfer. \finding{2}{Specialists share units.} With eight sources, 72 units form 44 groups; every multi-unit group crosses sources. Of retrieved units, $74\%$ come from other specialists, and one unit covers $94\%$ of gap queries (Sec.~\ref{sec:shared-distillation}). Shared code is $2.75\times$ the mean specialist increment and $0.34\times$ their concatenation, combining reusable physical structure across specialists.

\noindent{\textbf{Continual learning.}}
The eight-source harness adds two tasks per stage, querying the bank for unmet transitions. Fixed-target held-out SR rises from $51.3\%$ to $56.3\%$ after six additions and $75.0\%$ after eight, versus $53.8\%$ from scratch (Table~\ref{tab:ablation-distillation}B). Under the stopping rule, incremental target-stage offline tokens are lower than training from scratch ($106.20$\,M vs.\ $131.82$\,M), while total physical rollouts ($221$ vs.\ $100$ launches, including old-task retention checks) and combined tokens with eight-source distillation ($147.92$\,M) are higher (Supp.~\ref{sec:supp-paired-transfer}); across four stages, development SR stays at $88.1$-$100.0\%$ and the bank grows from 72 to 118 units.

\subsection{Ablation Study and Analysis}
\label{sec:ablation-cwl}
\label{sec:ablation-harness-optimization}
\noindent{\textbf{Harness optimization.}}
Table~\ref{tab:ablation-main} uses eight task chains, development conditions 11-13 (3/3 success or five rounds), and 80 held-out episodes on 1-10. \finding{1}{Beyond prompts.} Held-out SR rises from $47.50\%$ ($C$) to $52.50\%$ ($C{+}W$) and $55.00\%$ ($(C,W,L)$), though adding $L$ lowers development SR ($79.17\%\!\to\!75.00\%$; \challengeref{2}). \finding{2}{Reprojection drives gain.} Without reprojection, held-out SR falls to $47.50\%$ ($46.25\%$ without topology), despite more rounds and trials ($4.4, 13.6$). Fixed topology yields the lowest SR ($41.25\%$; Sec.~\ref{sec:emergent-topology}); graph updates without re-reading trials recover little. \finding{3}{Dynamic width reduces sequential rounds.} Serial search ($K{=}1$) uses the fewest trials ($11.4$) but requires $3.8$ sequential rounds. With parallel rollouts, dynamic $\kappa{=}30$ uses $5.5$ more trials for $2.9$ rounds at matched $55.00\%$ held-out SR ($83.33\%$ dev SR). Relative to dynamic width, fixed $K{=}3$ spends $11.2$ more trials for $+1.25$ points ($3.1$ rounds), while $K{=}2$ loses $2.50$.

\noindent{\textbf{Topology dynamics.}}
Over ten RoboDojo revisions, median transitions grow $4.0\times$ with required states unchanged (Fig.~\ref{fig:topology-read}). Under the final topology, $89\%$ of 159 failures retain progress (median $0.33$). Solved tasks rise $8\%\!\to\!52\%$ and mean best progress $0.46\!\to\!0.76$. With 78/149 revisions regressing, $58\%$ of 146 parent choices revisit earlier harnesses; re-read progress breaks development-SR ties. On LIBERO-PRO, transitions grow $2.2\times$, $79\%$ of failures retain progress, and $34\%$ of rounds reuse earlier parents. Across both benchmarks, re-reading failures exposes progress that a binary outcome alone would hide.

\noindent{\textbf{Scoring alignment.}}
Across 144 source-development executions on LIBERO-PRO (eight tasks, three initializations, $H_0$-$H_5$), independent simulator predicates resolve 48/57 physical conditions on 39 failures (Table~\ref{tab:simulator-alignment}). Physical P/R are $100\%$; progress MAE is $0$ on 30 fully resolved failures. The $29.17$-point gain in assigned physical progress corresponds to $25.00$-$32.64$ points under the reference. Final checkpoint regret is $[0,0.0524]$ for the full objective under the independent reference. Five fresh scoring passes on these traces select $H_5$ with zero progress SD (Supp.~\ref{sec:supp-method-implementation}).

\noindent{\textbf{Model sensitivity.}}
Table~\ref{tab:model-choices} varies one model role. \finding{1}{Roles decouple.} Online, Sonnet reaches $57.50\%$, Terra $55.00\%$, and DeepSeek $30.00\%$; offline, DeepSeek reaches $65.00\%$ and Sonnet $92.50\%$. Execution and optimization rankings differ. \finding{2}{Gains scale with base models.} With Terra execution fixed, Sonnet adds $37.50$ points over Sol through harness optimization without policy weight updates. At these settings, Sonnet gains $2.50$ points over Terra online versus $37.50$ over Sol offline. \methodname{} can thus translate stronger offline reasoning into better execution without retraining the policy or replacing the online executor.

\subsection{Case Study}
\label{sec:case-study}
\finding{1}{Reprojection changes old evidence.} In Fig.~\ref{fig:libero-case-study}a, $\Delta C$ separates lift from lateral motion. Tilted-plate success ($H_2$) makes supported deposit optional in $\Phi_2$; re-reading failed $H_1$ updates its acquisition witness $[70,140]\!\to\![70,90]$ and progress $0.50\!\to\!0.33$. \finding{2}{Retrieval guides edits.} The frontier retrieves rim/support and contact units from Spatial T04 and Object T05 (Fig.~\ref{fig:libero-case-study}b). A rim-pinch prompt ($\Delta C$) and seat-inspect-grasp workflow ($\Delta W$) place the bowl on held-out seed~4; matched Direct stalls after two picks.

\noindent{\textbf{Failure analysis.}} The MiniSet selects the hardest task in each category. With eight sources, transfer SR is $95\%$ on Spatial and Object but $10\%$ on Goal and $5\%$ on Long (Table~\ref{tab:ablation-distillation}A), reflecting the difficulty of these tasks. After eight task additions, continual learning reaches $75.0\%$ average target SR with greater total development effort (Table~\ref{tab:ablation-distillation}B), showing substantial progress on this challenging panel.

\subsection{Real-Robot Evaluation}
\label{sec:real-world}
On the dual-arm G2, we evaluate revisions $H_0$-$H_3$ with ten trials per task and version (Table~\ref{tab:real-world-tasks}; Fig.~\ref{fig:real-robot-sequences}). Sorting uses atomic actions, while pouring and plugging use a frozen VLA. With policy weights frozen throughout, sorting reaches $80\%$ SR from $20\%$ at $H_0$, with duration decreasing from $504.3$ to $402.1$\,s. Pouring improves from $40\%$ to $60\%$ and plugging from $50\%$ to $70\%$, with durations decreasing from $107.0$ to $95.4$\,s and $216.7$ to $175.4$\,s, respectively.

\begin{table}[t]
\begingroup
\input{tables/real_world_tasks}
\endgroup
\par\smallskip
\begingroup
\captionsetup{type=figure}
\centering
\includegraphics[width=\linewidth,trim=13bp 0 0 0,clip]{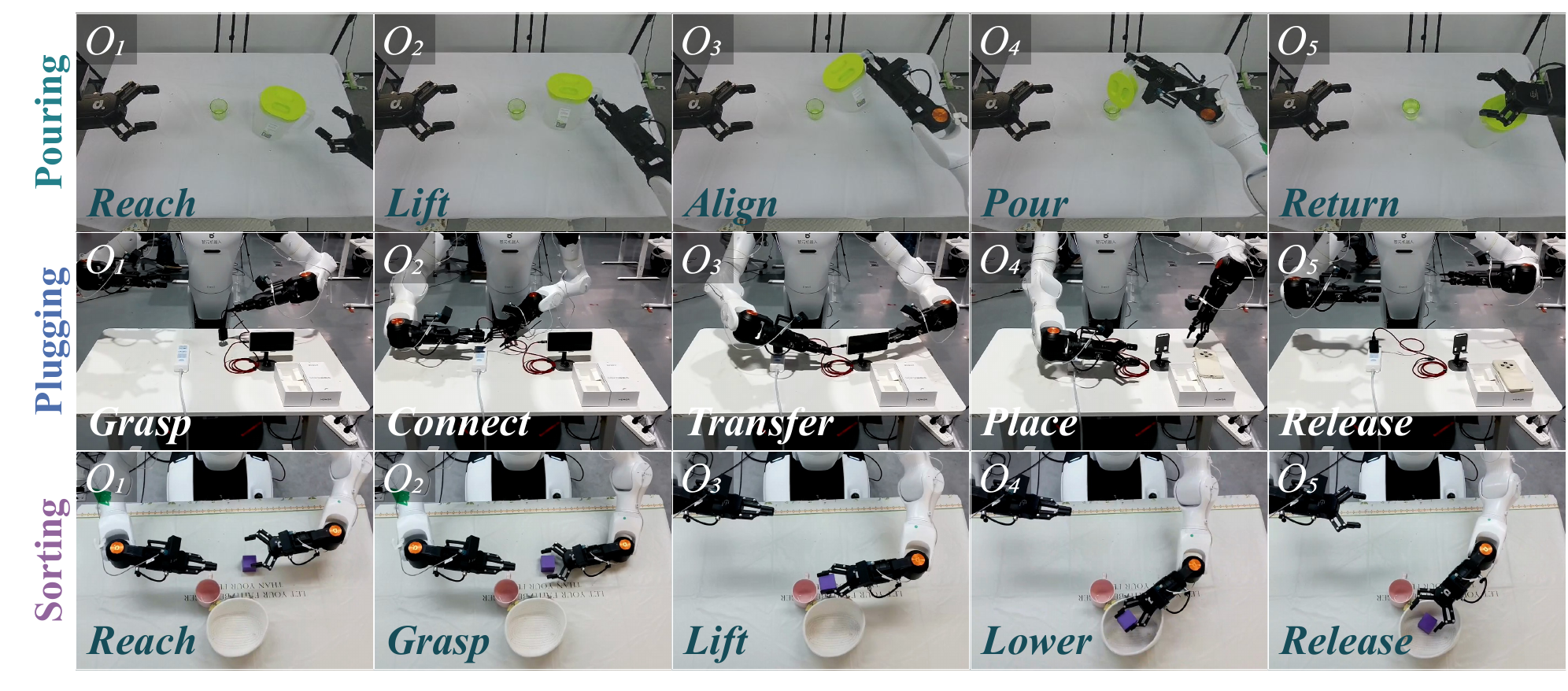}
\caption{\textbf{Real-robot task sequences.} Dual-arm G2 rollouts for pouring (VLA), plugging (VLA), and sorting (atomic actions).}
\label{fig:real-robot-sequences}
\endgroup
\end{table}
\par
\endgroup

%% file: tables/real_world_tasks.tex
\centering
\caption{\textbf{G2 evaluation.} Ten trials per task and version.}
\label{tab:real-world-tasks}
\footnotesize
\setlength{\tabcolsep}{0pt}
\setlength{\aboverulesep}{0pt}
\setlength{\belowrulesep}{0pt}
\renewcommand{\arraystretch}{1.06}
\begin{tabularx}{\linewidth}{@{}l *{6}{>{\centering\arraybackslash}X}@{}}
\toprule
\multicolumn{1}{c}{\multirow{2}{*}{\textbf{Config.}}} & \multicolumn{2}{c}{\textbf{Sorting}} & \multicolumn{2}{c}{\textbf{Pouring}} & \multicolumn{2}{c}{\textbf{Plugging}} \\
\cmidrule(lr){2-3}\cmidrule(lr){4-5}\cmidrule(lr){6-7}
 & \textbf{SR (\%)} & \textbf{Time (s)} & \textbf{SR (\%)} & \textbf{Time (s)} & \textbf{SR (\%)} & \textbf{Time (s)} \\
\midrule
$H_0$ & 20.0 & 504.3 & 40.0 & 107.0 & 50.0 & 216.7 \\
$H_1$ & \gainval{30.0}{10.0} & 472.6 & \gainval{50.0}{10.0} & 104.2 & \gainval{60.0}{10.0} & 204.8 \\
$H_2$ & \gainval{70.0}{50.0} & 425.4 & \gainval{50.0}{10.0} & 98.6 & \gainval{70.0}{20.0} & 188.2 \\
\rowcolor{tableaccent}
\methodlabel{$H_3$} & \gainval{\bestval{80.0}}{60.0} & \bestval{402.1} & \gainval{\bestval{60.0}}{20.0} & \bestval{95.4} & \gainval{\bestval{70.0}}{20.0} & \bestval{175.4} \\
\bottomrule
\end{tabularx}

%% file: sections/06_conclusion.tex
\section{Conclusion}
\label{sec:conclusion}

\methodname{} co-optimizes task understanding and an agent harness without weight updates. Transitions bind verified conditions, witnesses, and code; revised topologies re-score archived trials to guide optimization and transfer. Experiments show that this re-reading reveals progress in failed trials and supports reuse across tasks. The source code is publicly available.

\noindent{\textbf{Limitations.}}
Topology inference depends on observable traces, and ambiguous contacts can delay edge resolution. Optimization requires physical trials and model calls. Agent reasoning adds execution latency, limiting use in time-sensitive settings; transfer remains uneven across the hardest tasks in the initial eight-source setting.

%% file: sections/07_appendix.tex
\pdfbookmark[0]{Supplementary Material}{supplementary}
\section*{Supplementary Material}
\label{sec:appendix}

\newcommand{\suppstretch}{1.22}
\newcommand{\suppstyle}{\small\setlength{\tabcolsep}{8pt}%
  \setlength{\aboverulesep}{0pt}\setlength{\belowrulesep}{0pt}\renewcommand{\arraystretch}{\suppstretch}}
\newcommand{\suppth}[1]{\multicolumn{1}{c}{\textbf{#1}}}
\newcolumntype{Y}{>{\centering\arraybackslash}X}
\newcolumntype{Z}{>{\raggedright\arraybackslash}X}

\newcommand{\suppnav}[3][.8em]{%
  \noindent\hspace*{#1}\hyperref[#2]{\ref*{#2}\enspace #3}\nobreakspace\dotfill\pageref*{#2}\par}
\newtcolorbox{supppromptbox}[1]{unbreakable,enhanced jigsaw,colback=black!2.5,colframe=echoaccent!55!black,
  colbacktitle=echoaccent!70!black,coltitle=white,fonttitle=\small\bfseries\rmfamily,title={#1},
  boxrule=.55pt,arc=1.5pt,left=7pt,right=7pt,top=5pt,bottom=5pt,toptitle=2.5pt,bottomtitle=2.5pt,
  before skip=9pt,after skip=6pt,fontupper=\ttfamily\footnotesize\raggedright,parskip=4.5pt}
\newcommand{\suppprompt}[1]{#1\par\smallskip}

\begin{tcolorbox}[enhanced,colback=echoaccent!4,colframe=echoaccent!45,
  boxrule=0pt,leftrule=1.2pt,sharp corners,
  left=9pt,right=9pt,top=7pt,bottom=7pt,before skip=4pt,after skip=11pt]
{\sffamily\small\bfseries\color{echoaccent}Contents\par}
\vspace{4pt}
\small
\begin{minipage}[t]{.48\linewidth}
\suppnav[0pt]{sec:supp-method-implementation}{\textbf{Method and implementation}}
\suppnav{sec:supp-harness-implementation}{Harness components and interfaces}
\suppnav{sec:supp-topology-implementation}{Task topology and trajectory reprojection}
\suppnav{sec:supp-width-implementation}{Optimization and search width}
\suppnav{sec:supp-transfer-implementation}{Local topological units and meta-distillation}
\suppnav{sec:supp-simulator-alignment}{Simulator grounding and scoring stability}
\suppnav{sec:supp-prompts}{Prompt roles and examples}
\smallskip
\suppnav[0pt]{sec:supp-experimental-setup}{\textbf{Experimental setup}}
\suppnav{sec:supp-benchmark-splits}{Benchmark tasks and data splits}
\suppnav{sec:supp-model-settings}{Reference model settings}
\suppnav{sec:supp-evaluation-budgets}{Development and evaluation budgets}
\suppnav{sec:supp-comparison-protocol}{Comparison protocol and observation access}
\end{minipage}\hfill
\begin{minipage}[t]{.48\linewidth}
\suppnav[0pt]{sec:supp-additional-results}{\textbf{Additional results}}
\suppnav{sec:supp-libero-results}{LIBERO-PRO per-task results}
\suppnav{sec:supp-robodojo-results}{RoboDojo per-task results}
\suppnav{sec:supp-paired-transfer}{Paired transfer, continual learning, and costs}
\suppnav{sec:supp-archive-replay}{How retrospective reprojection guides search}
\suppnav{sec:supp-rerun}{Ablation stability on LIBERO-PRO}
\smallskip
\suppnav[0pt]{sec:supp-real-robot}{\textbf{Real-robot protocol}}
\end{minipage}
\end{tcolorbox}

\section{Method and Implementation Details}
\label{sec:supp-method-implementation}

\subsection{Harness Components and Interfaces}
\label{sec:supp-harness-implementation}
While the inner agent executes with a frozen foundation policy, the outer agent inspects logged observations, actions, and harness code between episodes to revise $(C,W,L)$. Within this division of responsibilities, context ($C$) filters and formats what the policy observes, workflow ($W$) implements bounded skill routines, and the control loop ($L$) governs observation timing, action batching, and replanning triggers. After each candidate edit, the host validates runtime interfaces, scores completed rollouts, and schedules the next parent and search width. Table~\ref{tab:supp-embodiment-h0} summarizes the frozen embodiment interfaces and seed harness $H_0=(C_0,W_0,L_0)$ across all three platforms.

\begin{table}[!htb]
\centering
\caption{\textbf{Frozen embodiment interfaces and seed harness $H_0=(C_0,W_0,L_0)$.} Within each optimization run, execution VLA weights and SAM3 perception remain fixed while optimization revises only $(C,W,L)$.}
\label{tab:supp-embodiment-h0}
\suppstyle
\setlength{\tabcolsep}{5pt}
\begin{tabularx}{\linewidth}{l Z Z Z}
\toprule
\textbf{Platform} & \textbf{Frozen policy \& primitives} & \textbf{Frozen perception} & \textbf{Seed harness $H_0$} \\
\midrule
LIBERO-PRO & Frozen $\pi_{0.5}$ VLA; Cartesian/gripper primitives (\texttt{move\_to}, \texttt{pi0\_pick}, \texttt{release}) & Two-camera RGB-D, frozen SAM3 (\texttt{segment}), calibrated depth (\texttt{back\_project}) & $C_0$: forwards latest images, proprioception, and task text; $W_0$: raw primitive wrappers; $L_0$: step-by-step observe--decide--act loop \\
RoboDojo & Frozen $\pi_{0.5}$ VLA; dual-arm IK and joint primitives (\texttt{move\_to}, \texttt{move\_pose}, \texttt{move\_joints}, \texttt{set\_gripper}, \texttt{pi0\_pick}) & Three-camera RGB-D, frozen SAM3 (\texttt{segment}), calibrated depth (\texttt{back\_project}) & Same task-agnostic $(C_0,W_0,L_0)$ interface across all 42 tasks \\
Real-world G2 & TCP atomic actions (Sorting) or frozen task VLA (\texttt{vla\_act}; Pouring, Plugging) & Head ($640{\times}400$) + dual wrist ($1280{\times}1056$ native / $640{\times}528$ planner) RGB-D, frozen SAM3 & Same task-agnostic $(C_0,W_0,L_0)$ interface across all three tasks \\
\bottomrule
\end{tabularx}
\end{table}

\subsection{Task Topology and Trajectory Reprojection}
\label{sec:supp-topology-implementation}
After each round, the analyst updates the task topology and reprojects every archived trajectory onto the revised graph, annotating which required milestones remained satisfied at termination, which were achieved and later lost, and where execution stalled. The host validates graph schema, interval coverage over the recorded frames, temporal continuity, and evidence pointers before computing retained progress, regression loss, and unmet frontiers (Eqs.~\ref{eq:retained-progress}--\ref{eq:reprojection-frontier}). Intervals marked \texttt{unknown} remain unassigned in the raw trace and contribute zero credit to the supported lower bound $\bar p_t^-$, and terminal goal assignments are restricted to officially successful rollouts.

\begin{figure}[H]
\centering
\includegraphics[width=\linewidth]{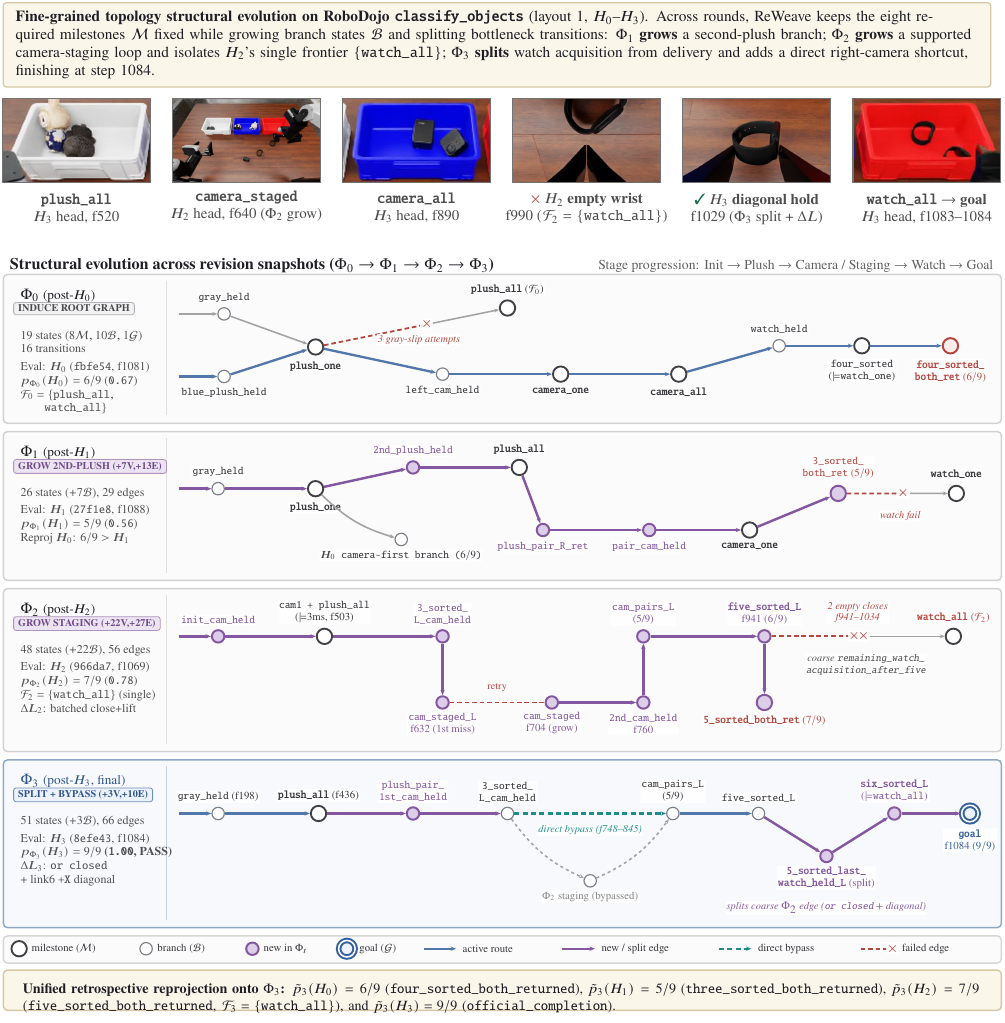}
\caption{\textbf{Topology evolution and trajectory reprojection on RoboDojo \texttt{classify\_objects}, single development layout (from the 25-task Fig.~\ref{fig:topology-read} topology-dynamics archive, separate from the 42-task Table~\ref{tab:robodojo-results} main run).} Each band shows the revised graph after one round ($H_0$ baseline and three updates $H_1$--$H_3$). Following two empty wrist closures in $H_2$ (frame~990), $H_3$ adopted a diagonal wrist approach and replanned upon jaw closure, completing the sort at step~1084. Table~\ref{tab:supp-topology-evolution} details each edit.}
\label{fig:supp-topology}
\end{figure}

\begin{table}[H]
\centering
\caption{\textbf{Graph revisions on \texttt{classify\_objects}.} Reprojected progress reflects re-evaluation onto that round's graph (denominator~9: eight required milestones plus terminal goal), and the final column pairs observed bottlenecks with the subsequent $(C,W,L)$ edits.}
\label{tab:supp-topology-evolution}
\suppstyle
\setlength{\tabcolsep}{4.5pt}
\begin{tabularx}{\linewidth}{@{}l c c Z Z@{}}
\toprule
\textbf{Checkpoint} & \suppth{Nodes / edges} & \suppth{Reprojected progress} & \textbf{Graph change and evidence} & \textbf{Frontier and edit} \\
\midrule
$H_0$ (\texttt{fbfe54}) & 19 / 16 & $6/9$ &
Four sorted objects at step 1081; three downward grasps failed on the second plush. &
\texttt{plush\_all} and \texttt{watch\_all} remained unmet; revise rim clearance for the plush ($C$). \\
\midrule
$H_1$ (\texttt{27f1e8}) & 26 / 29 & $6/9,\ 5/9$ &
\textbf{Grow.} Clearing the bin rim introduced \texttt{second\_plush\_held} between \texttt{plush\_one} and \texttt{plush\_all}. &
Execution stopped before the watches (step 1088); camera and watch frontiers motivated camera-first staging ($C$). \\
\midrule
$H_2$ (\texttt{966da7}) & 48 / 56 & $6/9,\ 5/9,\ 7/9$ &
\textbf{Grow.} A three-node branch staged and re-grasped the camera (frame 640); five objects sorted by step 1069. &
Two empty wrist closures (frames 941--1034) left only \texttt{watch\_all}; adopt a diagonal wrist approach ($C$) and replan after closure ($L$). \\
\midrule
\rowcolor{tableaccent}
$H_3$ (\texttt{8efe43}) & 51 / 66 & $6/9,\ 5/9,\ 7/9,\ \mathbf{9/9}$ &
\textbf{Split and bypass.} Two watch-grasp states replaced one coarse transition; direct camera delivery bypassed staging (steps 748--845). &
The new \texttt{or closed} guard returned control after jaw closure, completing the task at step 1084. \\
\bottomrule
\end{tabularx}
\end{table}

\begin{samepage}
Figure~\ref{fig:libero-case-study}(a) illustrates how newly acquired evidence revises historical credit: on Goal P05 (seed~12, from the independent Full optimization run in Sec.~\ref{sec:supp-rerun}), the $H_2$ prompt introduced a grasp check prior to lateral transport, and $H_2$ subsequently succeeded with the plate still tilted. Because that outcome showed that a supported intermediate deposit was not mandatory, the analyst reclassified that node as optional, reducing the earlier $H_1$ failure's retained progress from $0.50$ to $0.33$.\par
\end{samepage}

\subsection{Optimization and Search Width}
\label{sec:supp-width-implementation}
Each optimization round branches from an archived parent harness and evaluates $K_t$ parallel edits across all $N$ development conditions, whereas fixed-width baselines hold $K_t = K$ constant.

\begin{itemize}[leftmargin=1.6em,itemsep=3pt,topsep=4pt]
\item \textbf{Parent selection (Eq.~\ref{eq:parent-selection}).}
Candidates are ranked first by confirmed official development success and then by mean retained progress under $\Phi_t$. At equal success, an older version supersedes the best candidate from the latest round only if its progress lead reaches at least $0.10$; exact ties prefer the newer version.

\item \textbf{Forecast to Beta prior (Eq.~\ref{eq:width-prior}).}
For a parent solving $s$ of $N$ conditions ($g=N-s$ remaining), the analyst forecasts the probability $q_t$ that a single mutation solves all $g$ unmet conditions. Modeling conditional success probability as $\theta\sim\mathrm{Beta}(n_0m_t,n_0(1-m_t))$, the host solves for $m_t\in[0,1]$ via bisection such that $\mathbb E[\theta^g]=\prod_{i=0}^{g-1}(n_0m_t+i)/(n_0+i)=q_t$.

\item \textbf{Bellman width allocation (Eq.~\ref{eq:search-width}).}
Evaluating $K$ candidates incurs cost $K$ and yields gains $G_k\sim\mathrm{Binomial}(g,\theta)$, with $b=\max_k G_k$ and $S=\sum_kG_k$. Solving all remaining conditions ($b=g$) earns terminal reward $\kappa$; otherwise the posterior updates to $(s+b,\alpha+S,\beta+Kg-S)$ with one fewer round remaining. The host selects the smallest maximizing $K$ with $J_0=0$. This $s+b$ transition model is used solely for width allocation, while actual candidate evaluation scores all $N$ development conditions.
\end{itemize}

\begin{table}[!htb]
\centering
\caption{\textbf{Optimization and meta-distillation hyperparameters.} Prior concentration $n_0$ and completion value $\kappa$ were calibrated on pilot development trajectories.}
\label{tab:supp-hyperparameters}
\suppstyle
\setlength{\tabcolsep}{6pt}
\begin{tabularx}{\linewidth}{l c Z}
\toprule
\textbf{Parameter} & \suppth{Value} & \textbf{Role and calibration / definition} \\
\midrule
Rollback margin & $0.10$ & Minimum retained-progress gain needed for an older round to supersede the latest round at tied dev SR (exact ties prefer newest). \\
Prior strength $n_0$ & $3.0$ & Beta concentration in Eq.~\ref{eq:width-prior} (sibling correlation $\rho = 1/(n_0{+}1) = 0.25$); Beta-Binomial MLE over 150 pilot dev candidates across 79 rounds (flat over $n_0\in[2,4]$). \\
Completion value $\kappa$ & $30$ & Terminal reward in Eq.~\ref{eq:search-width}; calibrated on pilot dev forecast--outcome histories to balance search spend against development completion. \\
Width \& round caps & $K_{\max}{=}3,\ T_{\max}{=}5$ & Parallel candidate cap and MiniSet round limit ($T_{\max}{=}10$ with $K_t{=}1$ on single-seed main benchmarks). \\
Distillation weights & $\lambda_T{=}1.0,\ \lambda_S{=}10^{-3}\,\text{kB}^{-1}$ & Topological progress and code-excess weights in Eq.~\ref{eq:shared-distillation}, with reference threshold $b_{\mathcal B}{=}24.57\,\text{kB}$ (mean $(C,W,L)$ UTF-8 delta of eight source specialists over $H_0$). \\
Distillation stopping & $\tau{=}0.10,\ R_{\max}{=}5$ & Meta-loss stopping threshold and update round budget in Algorithm~\ref{alg:distillation}. \\
\bottomrule
\end{tabularx}
\end{table}

\subsection{Local Topological Units and Meta-Distillation}
\label{sec:supp-transfer-implementation}
For each student frontier gap $D_t(q,c)$, the host aligns source unit effects $E_u$ against missing target conditions under an injective object mapping $\sigma$ (Eq.~\eqref{eq:topology-alignment}). Each retrieved unit is a trajectory-supported mechanism binding whose entry and exit conditions were observed in archived rollouts; matching operates over object--property constraints via three-valued logic (set inclusion supports a condition, disjoint values contradict it, and unresolved relations remain unknown). The editor receives the covered target effects, source entry conditions, and links to the underlying code and witness traces, and chooses whether to adapt, compose, or decline each candidate before subsequent development rollouts validate the resulting harness.

\noindent\textbf{Direct distillation baseline.} Both editors in Table~\ref{tab:ablation-distillation}A access the same eight specialists' $(C,W,L)$ code, raw development records, and student rollout feedback, and share the same fixed source graphs and loss (Eq.~\eqref{eq:shared-distillation}). Direct (\emph{w/o topology}) omits only gap-indexed unit retrieval and evidence bindings. In Table~\ref{tab:ablation-distillation}A, the \emph{Selected $H$} SR column reports held-out source-task SR, whereas Prog.\ and Loss reflect the 24 source-development conditions (Table~\ref{tab:alignment-progress}).

\noindent\textbf{Unit consolidation and bowl placement.} Across the eight source specialists, the 72 trace-supported units group by shared object--property effects into 44 mechanism groups, with every multi-unit group spanning multiple source tasks. In Figure~\ref{fig:libero-case-study}(b), the top-left frame shows $H_1$ on source development (Spatial T04, seed~13) after locating the bowl prior to grasping. Gap retrieval surfaced \texttt{initial\_rim\_pick} (Spatial T04, rim and support constraints) and \texttt{metric\_initial\_pick} (Object T05, open-hand access), guiding an open-jaw inspection before closure in $H_2$; both methods selected $H_5$ on source development loss and were evaluated frozen on held-out Spatial P03 (seed~4).

\noindent\textbf{Continual-learning selection.} Table~\ref{tab:ablation-distillation}A selects checkpoints via Eq.~\eqref{eq:shared-distillation}. In Table~\ref{tab:ablation-distillation}B, task topologies guide Stage-1 frontier queries and mechanism retrieval while new-task graphs refine across rounds; to compare candidate checkpoints---including the incoming harness $R_0$---on a fixed scale as new-task topologies expand, checkpoint selection uses the mean development failure rate over all tasks seen so far plus the code-growth penalty ($0.001\,\text{kB}^{-1}$ above the incoming harness), while retaining topological progress as a diagnostic metric (see Sec.~\ref{sec:supp-paired-transfer} for stage selections, retention checks, and development costs).

\subsection{Simulator Grounding and Scoring Stability}
\label{sec:supp-simulator-alignment}
\noindent\textbf{Audit panel.} Six checkpoints ($H_0$--$H_5$) share the same 24 initializations across eight LIBERO-PRO source tasks (seeds~11--13), totaling 105 official successes, 39 failures, and 36,661 control steps. In this offline diagnostic cohort, archived LIBERO-PRO development records contain recorded simulator state alongside visual observations (whereas online execution and all RoboDojo runs, both online and offline, receive no simulator-state channel). Independent physical reference predicates (Table~\ref{tab:supp-reference-rules}), oracle labels, and counterfactual losses are computed post hoc solely for evaluation and withheld from both the scorer and optimizer.

\begin{table}[!htb]
\centering
\caption{\textbf{Physical reference rules} applied to the archived traces.}
\label{tab:supp-reference-rules}
\suppstyle
\begin{tabularx}{\linewidth}{l Z}
\toprule
\textbf{Condition} & \textbf{Independent reference} \\
\midrule
Placement & Official \texttt{in}/\texttt{on} predicate with the designated object binding. \\
Pickup & Object-specific grasp and at least $2$\,cm above initial support for three consecutive observations. \\
Retained control & Terminal control or a verified destination support. Pickup history may persist after release. \\
Drawer access & Closed or fully open joint endpoint within $1$\,mm. Intermediate positions remain unresolved. \\
\bottomrule
\end{tabularx}
\end{table}

\noindent\textbf{Resolution and accuracy.} Across the 39 failed trajectories, 57 physical and 21 epistemic target-grounding propositions arise. The reference predicates resolve 48 physical propositions, leaving nine physical propositions (such as intermediate drawer states) and all 21 epistemic propositions unresolved (Table~\ref{tab:alignment-states}). Precision and recall are computed over all 48 resolved physical propositions, and progress MAE is evaluated over the 30 failures with complete physical references.

\begin{table}[!htb]
\input{tables/simulator_alignment_states}
\end{table}

\noindent\textbf{Progress bounds.} A failure with $M$ necessary milestones, $a$ verified achievements, and $u$ unresolved milestones receives reference progress bounds $[a/(M+1),(a+u)/(M+1)]$, while official successes receive $1$. Physical progress applies the same interval bound to physical milestones alone.

\begin{figure}[!htb]
\centering
\includegraphics[width=.74\linewidth]{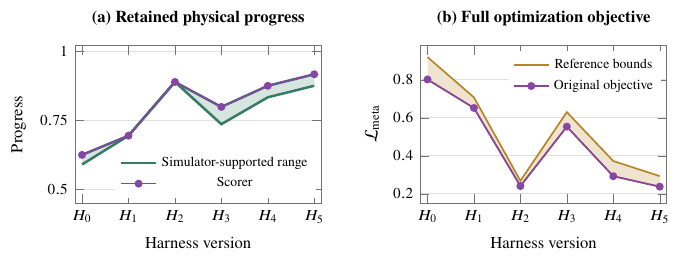}
\caption{\textbf{Scoring alignment across six checkpoints.} Shaded bands indicate independent simulator reference bounds for physical progress (a) and the full optimization objective (b).}
\label{fig:physical-alignment}
\end{figure}

\begin{table}[!htb]
\renewcommand{\suppstretch}{1.07}
\input{tables/simulator_alignment_progress}
\end{table}

\noindent\textbf{Checkpoint agreement.} From $H_0$ to $H_5$, assigned physical progress increases by 29.17 points (reference range 25.00--32.64) and full progress by 23.26 points (17.71--35.07). Replacing the progress term in Eq.~\eqref{eq:shared-distillation} with its reference bounds while holding official SR, code cost, and weights fixed yields the selection regret bounds in Tables~\ref{tab:alignment-progress} and~\ref{tab:alignment-selection}.

\begin{table}[!htb]
\renewcommand{\suppstretch}{1.07}
\input{tables/simulator_alignment_selection}
\end{table}

\noindent\textbf{Repeated scoring.} Rescoring the 39 failures across five independent passes of six fresh sessions each yields TP/FP/FN/TN counts of $16/0/0/32$, physical MAE $0.0000$ ($n=30$), a full-objective regret envelope of $[0,0.0524]$, and selection of $H_5$ across all five passes (zero progress SD across passes).

\noindent\textbf{Task-level uncertainty.} Bootstrapping the eight task clusters 10,000 times (keeping versions, initializations, and scoring passes intact) yields 95\% physical-progress gain bounds of $[1.39,55.56]$ points and regret bounds of $[0,0.0839]$.

\subsection{Prompt Roles and Examples}
\label{sec:supp-prompts}
The excerpts below are taken from the offline instruction templates; ellipses mark omitted passages.

\begin{supppromptbox}{Topology analyst: forecast for the next round}
\suppprompt{Alongside the graph, supply forecast.outcomes, forecast.confidence (0-1), and forecast.rationale. outcomes is a finite distribution of \{progress, probability\} for ONE complete child candidate after ONE independent author modifies the Host-selected parent with the same available graph evidence, author time budget and episode limits. \ldots{} Each progress is the arithmetic mean TERMINAL graph progress (0-100) across the parent's full development conditions, equally weighted on THIS graph's scale. Include currently successful seeds and potential regressions. \ldots}

\suppprompt{Confidence audits your certainty in the evidence supporting this distribution; it is not a gain discount or a variance estimate. \ldots}
\end{supppromptbox}

\begin{supppromptbox}{Topology analyst: graph construction}
\suppprompt{Each new round starts a fresh native session. The previous COMPLETE graph, progress map and evidence index carry forward the run's analysis. \ldots}

\suppprompt{EVERY node, including every optional branch, must represent meaningful task progress that was not obtained for free. Name the positive condition acquired, explain how it enables or completes this task, and cite evidence of its acquisition through actual observation, grounding, tool use or control. \ldots}

\suppprompt{Node predecessors reference immediate MILESTONE NODE IDs in this same graph, whose achievements remain satisfied at this state. They form an acyclic task backbone. All predecessors are jointly satisfied, so alternate implementations are branches, not additional mandatory achievements. A milestone adds itself, a branch adds nothing, and each goal includes the entire milestone backbone. \ldots}

\suppprompt{Only officially successful final frames may use a goal node. Host computes effective progress from these graph relations, never from model-provided numbers.}
\end{supppromptbox}

\begin{supppromptbox}{Topology analyst: reprojection of archived trials}
\suppprompt{Reproject EVERY historical episode under this one graph. All original frame indices must be tiled exactly once. Use status=node with a node\_id only when its earned progress condition is supported. Use status=background and node\_id=null for known absence of retained task progress; it is zero progress, not a node or an achievement. Use status=unknown and node\_id=null with a reason when evidence is insufficient; Host preserves null progress, not an invented zero.}

\suppprompt{Revise earlier interpretations when evidence warrants. A later outcome can challenge an earlier interpretation, but never defines an earlier state as eventually successful. Contradicting successful routes require revision of the backbone and reprojection of all history. \ldots}

\suppprompt{Do not discard an already earned condition merely because the gripper is empty: retain the supported progress node while its task-relevant achievement still holds. \ldots}

\suppprompt{Each edge declares object roles, entry\_conditions and exit\_conditions. Entry includes relevant physical prerequisites and publicly available history; exit describes the acquired condition AND what matters for continuing afterwards. Do not label an entrance by a future success/failure. Interface conditions are evidence-backed interpretations, not proven universal pre/postconditions.}
\end{supppromptbox}

\begin{supppromptbox}{Harness author: one mutation}
\suppprompt{Improve ONE complete candidate under the objective and budgets in context/input.json. Read the actual task instructions and named current rules; historical verdicts retain their original criteria. Official success across the development conditions is the objective. Successes from different candidates cannot be combined into one candidate's score. Infrastructure-missing outcomes remain unknown.}

\suppprompt{\ldots{} Earlier explanations are revisable beliefs, not established causes.}

\suppprompt{Start from the observed execution problem and inspect the original model inputs, choices and feedback needed to understand it. For physical claims, inspect relevant video or images alongside the trace. Keep observed facts and causal hypotheses distinct; record a material unresolved ambiguity in plan.json. \ldots}

\suppprompt{Choose components by the supported mechanism, without an edit quota. Editable components are permissions, not required changes; a change to one component is sufficient.}

\suppprompt{Integrate the change into the existing implementation; remove superseded rules and conflicting patches. Preserve demonstrated behavior and its conditions, not all inherited code. Geometry and task bindings must be grounded in current public observations; no task/seed dispatch, copied action scripts or privileged online inputs.}
\end{supppromptbox}

\begin{supppromptbox}{Meta-distillation: shared harness from specialists}
\suppprompt{Construct ONE compact cross-task executable harness, not a library of complete task specialists. Read context/input.json and the admitted source evidence. Refactor coupled C/W/L mechanisms into common implementations; preserve only necessary branches on observable physical conditions. Do not dispatch on benchmark task IDs, memorized task names, reset IDs or evaluator outcomes. Do not embed whole teacher harnesses behind a router. You may discover common mechanisms yourself.}

\suppprompt{Delta size counts added/replacement UTF-8 source/text/schema lines against context/input.json's fixed size\_reference. Unchanged base content costs zero; removals never cancel new content. \ldots{} Reference topologies stay fixed; the Host independently projects new trajectories, with official success scoring 1 and necessary retained milestones providing partial credit. Optional branches and retries add no credit. \ldots}

\suppprompt{Ten-seed specialist degradation is measured only after the shared Harness freezes; never request those test outcomes or use them to revise a candidate. \ldots{} Fixtures check interfaces only; only subsequent Host physical evaluations establish task success.}
\end{supppromptbox}

\begin{supppromptbox}{Meta-distillation: topology-guided retrieval and continual update}
\suppprompt{When context/shared\_topology.json exists, it is a lossless source factorization: read BOTH common mechanisms AND every source residual/qualification/dependency. Do not replace source-specific guards with the intersection of common conditions. \ldots}

\suppprompt{When context/input.json provides gap\_index, begin with that CURRENT student's observed gaps. query\_student\_gap reads the compact mechanism index, candidate summaries, individual bindings and complete teacher-interface details. \ldots{} This is navigation, not top-k selection. Full and partial result matches are references, not a ranking or executable plan. Compare code and evidence yourself; choose useful references, combine them, or decline them. \ldots}

\suppprompt{Repair the preceding student while preserving protected\_outcomes; avoid rebuilding unrelated working mechanisms. \ldots{} Do not claim success from a retrieval match. Overall official\_success is never an operation correspondence. No held-out feedback is allowed.}

\suppprompt{Continual updates also use the positive part of size excess; compression never offsets task failure. In continual mode, update ONE shared C/W/L directly on all declared old-task and new-task development conditions. Do not train separate new task specialists or add a later merge stage. Old-task topologies are immutable; the Host refreshes the two new-task topologies from each evaluated shared version. Their current frontiers query only the old-task mechanism bank.}
\end{supppromptbox}

\section{Experimental Setup}
\label{sec:supp-experimental-setup}

\subsection{Benchmark Tasks and Data Splits}
\label{sec:supp-benchmark-splits}
Tables~\ref{tab:supp-benchmark-splits} and~\ref{tab:supp-miniset-panels} summarize the development and held-out test splits across all benchmarks. In zero-shot transfer, target tasks remain unseen until held-out evaluation; in continual learning, target pairs enter development sequentially while held-out seeds~1--10 remain untouched. For RoboDojo, Generalization averages five standard and five randomized test layouts per task, while the remaining capabilities use ten standard layouts. Fig.~\ref{fig:libero-case-study}(a) uses the independent Full rerun.

\begin{table}[H]
\centering
\caption{\textbf{Development and held-out panels} used in the main paper.}
\label{tab:supp-benchmark-splits}
\renewcommand{\suppstretch}{1.10}\suppstyle
\begin{tabularx}{\linewidth}{Z Z Z}
\toprule
\textbf{Experiment} & \textbf{Development panel} & \textbf{Held-out panel} \\
\midrule
LIBERO-PRO (Table~\ref{tab:main-results}) & 80 tasks, seed~0 per task & 80 tasks, seeds~1--10 per task \\
RoboDojo (Table~\ref{tab:robodojo-results}) & 42 tasks, seed~0 (one standard layout) per task & 42 tasks, seeds~1--10 per task \\
Eight-source transfer (Table~\ref{tab:ablation-distillation}A) & Eight source tasks, seeds~11--13 & Eight separate target tasks, seeds~1--10 \\
Continual learning (Table~\ref{tab:ablation-distillation}B) & Two target tasks added per stage, seeds~11--13 & Fixed eight-task target panel, seeds~1--10 \\
MiniSet ablation (Table~\ref{tab:ablation-main}) & Eight tasks listed in Table~\ref{tab:supp-rerun}, seeds~11--13 & Same tasks, seeds~1--10 \\
\bottomrule
\end{tabularx}
\end{table}

\begin{table}[H]
\centering
\caption{\textbf{LIBERO-PRO MiniSet task IDs.} Each stage adds two tasks, pairing one position-perturbed (\texttt{P}) and one task-perturbed (\texttt{T}) task per suite. Table~\ref{tab:ablation-main} uses the target panel with Spatial \texttt{T04} replacing Object \texttt{P03}.}
\label{tab:supp-miniset-panels}
\renewcommand{\suppstretch}{1.10}\suppstyle
\setlength{\tabcolsep}{6pt}
\begin{tabularx}{\linewidth}{l c Z Z}
\toprule
\textbf{Stage} & \suppth{Suite} & \textbf{Source (Table~\ref{tab:ablation-distillation}A, Sec.~\ref{sec:supp-simulator-alignment})} & \textbf{Target (Table~\ref{tab:ablation-distillation}A--B)} \\
\midrule
Stage~1 ($2$ / $+2$) & Spatial & \texttt{P01}, \texttt{T04} & \texttt{P03}, \texttt{T08} \\
Stage~2 ($4$ / $+4$) & Object & \texttt{P08}, \texttt{T05} & \texttt{P03}, \texttt{T07} \\
Stage~3 ($6$ / $+6$) & Goal & \texttt{P03}, \texttt{T03} & \texttt{P05}, \texttt{T00} \\
Stage~4 ($8$ / $+8$) & Long & \texttt{P05}, \texttt{T00} & \texttt{P04}, \texttt{T06} \\
\bottomrule
\end{tabularx}
\end{table}

\subsection{Reference Model Settings}
\label{sec:supp-model-settings}
Table~\ref{tab:supp-model-settings} lists the default online controller and offline optimizer models for each benchmark panel; Table~\ref{tab:model-choices} in the main paper varies one role at a time relative to the LIBERO-PRO MiniSet reference.

\begin{table}[H]
\centering
\caption{\textbf{Reference model settings.} Default online LLM controller and offline optimizer models per benchmark panel; within each run, execution VLA weights and SAM3 stay fixed.}
\label{tab:supp-model-settings}
\renewcommand{\suppstretch}{1.10}\suppstyle
\begin{tabularx}{\linewidth}{l Z Z}
\toprule
\textbf{Panel} & \textbf{Online controller} & \textbf{Offline optimization} \\
\midrule
LIBERO-PRO main & GPT-6 Astra (low) & GPT-6.1 Sol (xhigh) \\
RoboDojo main & GPT-6.1 Sol (xhigh) & GPT-6.1 Sol (xhigh) \\
LIBERO-PRO MiniSet reference & GPT-6.1 Terra (high) & GPT-6.1 Sol (xhigh) \\
\bottomrule
\end{tabularx}
\end{table}

\subsection{Development and Evaluation Budgets}
\label{sec:supp-evaluation-budgets}
Table~\ref{tab:supp-evaluation-budgets} compares development scopes, physical step horizons, and agent turn limits alongside published LIBERO-PRO presets. LIBERO-PRO limits include ten settling steps, whereas RoboDojo enforces each task's native simulator horizon at reset. A single controller decision may invoke a multi-step workflow, and decision counts exclude calls to the frozen VLA and SAM3. Both main benchmarks develop with search width $K_t = 1$.

\begin{table}[H]
\centering
\caption{\textbf{Development and execution budgets.} External entries reflect published LIBERO-PRO presets (cited from their respective papers); NR denotes an unreported limit. MiniSet ablations (Table~\ref{tab:ablation-main}) use the hardest eight-task subset at $1\times$ horizons (where $H_0$ scores $3.75\%$ vs.\ $89.00\%$ on the full 80-task Table~\ref{tab:main-results} benchmark).}
\label{tab:supp-evaluation-budgets}
\renewcommand{\suppstretch}{1.10}\suppstyle
\setlength{\tabcolsep}{4pt}
\begin{tabularx}{\linewidth}{l Z c c c}
\toprule
\textbf{Method} & \textbf{Task-local development} & \suppth{Test states} & \suppth{Sim.\ steps} & \suppth{Agent budget} \\
\midrule
ORA~\cite{ora2026code} & No task-local search & Official states & 10,000 & 80 robot / 200 model calls \\
OpenRUA~\cite{chu2026openrua} & Fresh workspace each episode & Indices 0--9 & Unlimited & 500 turns \\
RPent~\cite{zhang2026harness} & Task memory on seed 0 & Seeds 1--10 & 10,000 & 100 turns \\
ASPIRE~\cite{lu2026aspire} & Repair and search on seeds 51--65 & Seeds 1--50 & NR & NR \\
\midrule
LIBERO-PRO $H_0$ & None & Seeds 1--10 & 230--2,610 ($1\times$--$5\times$) & 100 decisions \\
\rowcolor{tableaccent}
LIBERO-PRO, \methodname{} & Seed 0, up to ten attempts & Seeds 1--10 & 230--2,610 ($1\times$--$5\times$) & 100 decisions \\
\rowcolor{tableaccent}
RoboDojo, \methodname{} & Seed 0, up to ten attempts & Seeds 1--10 & 200--1,900 (official $1\times$) & 100 decisions \\
\bottomrule
\end{tabularx}
\end{table}

\subsection{Comparison Protocol and Observation Access}
\label{sec:supp-comparison-protocol}
Table~\ref{tab:supp-comparison-controls} details the evaluation protocol, observation boundaries, and development isolation enforced across experiments.

\begin{table}[H]
\centering
\caption{\textbf{Comparison protocols and observation access.}}
\label{tab:supp-comparison-controls}
\renewcommand{\suppstretch}{1.10}\suppstyle
\setlength{\tabcolsep}{6pt}
\begin{tabularx}{\linewidth}{l Z}
\toprule
\textbf{Dimension} & \textbf{Protocol} \\
\midrule
External systems & ASPIRE, RPent, and HexaShell entries are taken from published results under their respective adaptation procedures~\cite{lu2026aspire,zhang2026harness,gao2026hexashell}. \\
Online observations & Both $H_0$ and \methodname{} access RGB-D streams, calibration, proprioception, task language, object-name labels, and SAM3 depth back-projection; ground-truth simulator poses, masks, forces, and rewards are excluded. \\
Offline traces & Archived LIBERO-PRO development records contain simulator-state telemetry used to compute the independent post-hoc reference predicates (Sec.~\ref{sec:supp-simulator-alignment}), which are withheld from the scorer and optimizer; RoboDojo has no simulator-state channel either online or offline. \\
Development isolation & Main runs adapt on seed~0 for up to ten rounds under task-agnostic observation interfaces (without task/seed dispatch or action-script replay) and freeze $(C,W,L)$ prior to testing on seeds~1--10. \\
MiniSet accounting & MiniSet runs develop on seeds~11--13 at $1\times$ horizons; Sec.~\ref{sec:supp-paired-transfer} breaks down cross-task distillation costs and continual old-task regression checks. \\
\bottomrule
\end{tabularx}
\end{table}

\clearpage
\section{Additional Results}
\label{sec:supp-additional-results}

\subsection{Detailed LIBERO-PRO Results}
\label{sec:supp-libero-results}
Table~\ref{tab:supp-libero-per-task} reports per-task held-out success rates and online execution costs across all 80 LIBERO-PRO tasks (790 successes over 800 task--seed evaluations, $98.75\%$), using the model settings in Table~\ref{tab:supp-model-settings}.

\input{tables/libero_per_task}

\clearpage
\subsection{Detailed RoboDojo Results}
\label{sec:supp-robodojo-results}
Table~\ref{tab:supp-robodojo-per-task} reports all 42 RoboDojo tasks on held-out seeds~1--10 under native simulator step limits, with a five-capability macro-average of $63.43$ Score and $58.25\%$ SR. The multi-round traces in Figs.~\ref{fig:topology-read} and~\ref{fig:supp-topology}, Table~\ref{tab:supp-topology-evolution}, and Sec.~\ref{sec:supp-archive-replay} come from the separate 25-task topology-dynamics archive, so per-task update counts differ from Table~\ref{tab:supp-robodojo-per-task}.

\input{tables/robodojo_per_task}

\clearpage
\subsection{Paired Transfer, Continual Learning, and Cost Accounting}
\label{sec:supp-paired-transfer}
On the eight unseen target tasks in Table~\ref{tab:ablation-distillation}A, \methodname{} solves 41 of 80 held-out episodes versus 33 for Direct distillation (Table~\ref{tab:supp-paired-transfer}). Across matched initializations, \methodname{} wins 11 pairs where Direct fails and loses 3, yielding net gains of $+7$ on Spatial, $0$ on Object, $+2$ on Goal, and $-1$ on Long.

\begin{table}[!htb]
\centering
\caption{\textbf{Eight-source transfer.} Held-out successes across ten matched initializations per task. P and T denote position and task perturbations; $\Delta$SR reports the gain over Direct (pp).}
\label{tab:supp-paired-transfer}
\suppstyle
\setlength{\tabcolsep}{3pt}
\renewcommand{\arraystretch}{1.15}
\begin{tabularx}{\linewidth}{l *{7}{Y}}
\toprule
 & \multicolumn{2}{c}{\textbf{Successful episodes}} & \multicolumn{4}{c}{\textbf{Paired outcomes}} & \\
\cmidrule(lr){2-3}\cmidrule(lr){4-7}
\textbf{Held-out task} & \suppth{\methodname{}} & \suppth{Direct} & \suppth{Both succeed} & \suppth{\methodname{} only} & \suppth{Direct only} & \suppth{Both fail} & \suppth{$\Delta$SR} \\
\midrule
Spatial P03 & 9 & 4 & 4 & 5 & 0 & 1 & $+50$ \\
Spatial T08 & 10 & 8 & 8 & 2 & 0 & 0 & $+20$ \\
Object P03 & 9 & 9 & 8 & 1 & 1 & 0 & $0$ \\
Object T07 & 10 & 10 & 10 & 0 & 0 & 0 & $0$ \\
Goal P05 & 2 & 0 & 0 & 2 & 0 & 8 & $+20$ \\
Goal T00 & 0 & 0 & 0 & 0 & 0 & 10 & $0$ \\
Long P04 & 0 & 1 & 0 & 0 & 1 & 9 & $-10$ \\
Long T06 & 1 & 1 & 0 & 1 & 1 & 8 & $0$ \\
\midrule
\rowcolor{tableaccent}
\textbf{Overall (80 pairs)} & \textbf{41} & \textbf{33} & \textbf{30} & \textbf{11} & \textbf{3} & \textbf{36} & $\mathbf{+10}$ \\
\bottomrule
\end{tabularx}
\end{table}

\noindent\textbf{Continual learning.} In Table~\ref{tab:ablation-distillation}B, the eight target tasks enter development in pairs while seeds~1--10 remain held out on the same 80-episode evaluation panel. Starting from the eight-source shared harness ($41/80$), the retained checkpoint stays unchanged through $+2$ and $+4$, improves to $45/80$ at $+6$, and reaches $60/80$ at $+8$---an adaptation gain of 17 episodes ($+21.25$\,pp, including $+8$ on Long tasks) over independent From Scratch training ($43/80$; Table~\ref{tab:supp-continual-targets}). At $+8$, task-macro development SR is $93.8\%$ (following $100.0\%$, $97.2\%$, and $88.1\%$ at $+2$, $+4$, and $+6$); per trial, $6/6$ new-task and $13/14$ earlier-task development checks pass.

\begin{table}[!htb]
\centering
\caption{\textbf{Continual learning.} Held-out success counts over ten seeds per task. Gain compares the $+8$ checkpoint against From Scratch.}
\label{tab:supp-continual-targets}
\suppstyle
\renewcommand{\arraystretch}{1.15}
\begin{tabularx}{\linewidth}{l *{5}{Y}}
\toprule
\textbf{Held-out task} & \suppth{Start, $+2$, $+4$} & \suppth{$+6$ tasks} & \suppth{From Scratch} & \suppth{$+8$ tasks} & \suppth{Gain} \\
\midrule
Spatial P03 & 9 & 9 & 1 & 9 & $+8$ \\
Spatial T08 & 10 & 10 & 10 & 9 & $-1$ \\
Object P03 & 9 & 10 & 10 & 10 & $0$ \\
Object T07 & 10 & 10 & 7 & 8 & $+1$ \\
Goal P05 & 2 & 2 & 2 & 3 & $+1$ \\
Goal T00 & 0 & 0 & 1 & 1 & $0$ \\
Long P04 & 0 & 0 & 7 & 10 & $+3$ \\
Long T06 & 1 & 4 & 5 & 10 & $+5$ \\
\midrule
\rowcolor{tableaccent}
\textbf{Overall (80 trials)} & \textbf{41} & \textbf{45} & \textbf{43} & \textbf{60} & $\mathbf{+17}$ \\
\bottomrule
\end{tabularx}
\end{table}

\noindent\textbf{Confidence intervals.} Bootstrapping matched initializations within tasks gives 95\% intervals of $[2.50,17.50]$\,pp over Direct (Table~\ref{tab:supp-paired-transfer}) and $[11.25,31.25]$\,pp over From Scratch (Table~\ref{tab:supp-continual-targets}). Cluster bootstrapping over intact tasks yields $[-1.25,23.75]$\,pp and $[3.75,42.50]$\,pp, respectively, while leaving out any single target task preserves gains of $4.3$ to $12.9$\,pp over Direct.

\noindent\textbf{Development cost accounting.} In eight-source distillation (Table~\ref{tab:ablation-distillation}A), both \methodname{} and Direct ran five updates over 144 scored development episodes (24 shared initial episodes plus five updates over $8\times 3$ source trials), with \methodname{} reaching $41/80$ vs.\ $33/80$ held-out successes using $41.72$\,M vs.\ $47.72$\,M offline tokens ($-12.6\%$; $120$ vs.\ $124$ post-initialization launches across $11$ vs.\ $10$ sessions).

\clearpage
\noindent\textbf{Table~\ref{tab:ablation-distillation}B accounting.} At $+2$ and $+4$, the incoming harness $R_0$ met the stopping rule ($\mathcal L^*<0.10$) and advanced directly. At $+4$, this cutoff gives six scored new-task records, eight launches, and $5.71$\,M tokens. The $+6$ and $+8$ stages selected $R_1$ and $R_4$, respectively.

\noindent{}Table~\ref{tab:supp-stage-costs} reports $106.20$\,M target-stage offline tokens under the stopping rule. Including eight-source distillation ($41.72$\,M) gives $147.92$\,M. From Scratch used $131.82$\,M across eight tasks: 105 scored records (16 cached), 100 launches, and 68 offline sessions. The main table shows the per-pair mean. Its loss $0.148$ combines the checkpoint selected independently for each task; $0.234$ is a round-aligned portfolio mean.

\begin{table}[H]
\centering
\caption{\textbf{Continual-learning development costs (Table~\ref{tab:ablation-distillation}B).} Shared-harness counts report stopping-rule target-stage costs (using the retained $R_0$ cutoff for Object $+4$) and include old-task checks and retries (unchanged incoming harnesses reuse cached old-task checks, so $+4$ at $R_0$ launches $2$ uncached checks). From Scratch totals all eight tasks (Table~\ref{tab:ablation-distillation}B reports the per-pair mean); totals are computed from unrounded values.}
\label{tab:supp-stage-costs}
\suppstyle
\setlength{\tabcolsep}{4pt}
\begin{tabularx}{\linewidth}{@{}l Y Y c Y Y Y Y@{}}
\toprule
 & \multicolumn{4}{c}{\textbf{Shared harness (\methodname{})}} & \multicolumn{3}{c}{\textbf{From Scratch}} \\
\cmidrule(lr){2-5}\cmidrule(lr){6-8}
\textbf{New task pair} & \suppth{Upd.} & \suppth{New rec.} & \multicolumn{1}{c}{\tablehead{\textbf{Launches}\\[-1pt]\textbf{(new+old)}}} & \suppth{Tokens (M)} & \suppth{New rec.} & \suppth{Launches} & \suppth{Tokens (M)} \\
\midrule
Spatial ($+2$) & 0 & 6 & 14 (6+8) & 5.10 & 27 & 28 & 33.45 \\
Object ($+4$), $R_0$ & 0 & 6 & 8 (6+2) & 5.71 & 12 & 9 & 13.71 \\
Goal ($+6$) & 5 & 36 & 99 (36+63) & 53.93 & 33 & 33 & 42.75 \\
Long ($+8$) & 4 & 30 & 100 (33+67) & 41.46 & 33 & 30 & 41.92 \\
\midrule
\rowcolor{tableaccent}
\textbf{All eight tasks} & \textbf{9} & \textbf{78} & \textbf{221 (81+140)} & \textbf{106.20} & \textbf{105} & \textbf{100} & \textbf{131.82} \\
\bottomrule
\end{tabularx}
\end{table}

\subsection{How Retrospective Reprojection Guides Search}
\label{sec:supp-archive-replay}
Reprojection updates the frontier, physical conditions, and witness frames for earlier failures. In Table~\ref{tab:ablation-main}B, both variants update the graph and can read old trajectories. Full recomputes their structured annotations; \emph{w/o reprojection} scores new trials on the current graph and keeps earlier scores, frontiers, and witnesses fixed.

\begin{itemize}[leftmargin=1.6em,itemsep=3pt,topsep=4pt]
\item \begin{samepage}
\textbf{Deposit coin.} The parent stayed fixed through round~8. Re-reading two wrist frames distinguished returning with the coin under control from returning empty-handed. Later, reassessing head frames~220 and~232 alongside the controller trace located a $51.7$\,mm alignment residual before the round~9 completion.\par
\end{samepage}
\item \begin{samepage}
\textbf{Sort nesting dolls by size.} With the same parent across rounds 0--3, reprojecting two earlier failures narrowed the frontier to ordered placement. One archived score rose from $0.375$ to $0.50$. Head frames~989 and~1026 exposed the row-alignment error before completion at frame~1038.\par
\end{samepage}
\end{itemize}

\begin{samepage}
\noindent\textbf{Parent-choice replay.} The $0.10$ margin applies at tied development success. Replaying Eq.~\eqref{eq:parent-selection} on the fixed 146-round RoboDojo development archive changed archived progress in 20 rounds and the selected parent in seven. In \emph{Pour balls into a vase}, the reprojected progress of $H_0$ rose from $0.25$ to $0.40$, above $H_1$ at $0.20$.\par
\end{samepage}

\clearpage
\subsection{Ablation Stability on LIBERO-PRO}
\label{sec:supp-rerun}
The independent Full rerun starts from the same archived $H_0$ traces, develops on seeds~11--13 for up to five updates, and evaluates seeds~1--10 under the $1\times$ horizon. It solves $43/80$ held-out episodes, compared with $44/80$ in Table~\ref{tab:ablation-main} (Table~\ref{tab:supp-rerun}).

\begin{table}[H]
\centering
\caption{\textbf{Independent Full rerun on LIBERO-PRO.} Per-task held-out success rates (seeds~1--10), development effort (after $H_0$), and test execution counts alongside main-paper Table~\ref{tab:ablation-main}.}
\label{tab:supp-rerun}
\renewcommand{\suppstretch}{1.10}\suppstyle
\setlength{\tabcolsep}{5pt}
\begin{tabularx}{\linewidth}{l *{5}{Y}}
\toprule
\textbf{Task} & \suppth{Held-out SR (\%)} & \suppth{Updates} & \suppth{Dev.\ trials} & \suppth{Test dec.} & \suppth{Test steps} \\
\midrule
Spatial P03 & 40.0 (4/10) & 4 & 12 & 21.2 & 206.7 \\
Spatial T04 & 30.0 (3/10) & 5 & 15 & 14.9 & 203.3 \\
Spatial T08 & 30.0 (3/10) & 1 & 3 & 10.3 & 214.6 \\
Object T07 & 100.0 (10/10) & 3 & 9 & 9.0 & 224.7 \\
Goal P05 & 40.0 (4/10) & 5 & 15 & 12.4 & 249.2 \\
Goal T00 & 40.0 (4/10) & 5 & 15 & 11.8 & 275.3 \\
Long P04 & 70.0 (7/10) & 2 & 7 & 25.6 & 415.7 \\
Long T06 & 80.0 (8/10) & 5 & 15 & 11.5 & 403.9 \\
\midrule
\rowcolor{tableaccent}
\textbf{Independent rerun (Full)} & \textbf{53.75 (43/80)} & \textbf{3.75} & \textbf{11.38} & \textbf{14.6} & \textbf{274.2} \\
\rowcolor{tableaccent!50}
\textbf{Main paper (Table~\ref{tab:ablation-main}, Full)} & 55.00 & 3.8 & 11.4 & 12.0 & 274.2 \\
\bottomrule
\end{tabularx}
\end{table}

\section{Real-Robot Protocol}
\label{sec:supp-real-robot}
On the dual-arm G2 robot, each checkpoint ($H_0$--$H_3$) was evaluated over ten physical trials per task on ten matched workspace layouts (Table~\ref{tab:real-world-tasks}). The harness was fixed within each batch of ten trials and revised offline between batches from archived recordings while policy weights remained fixed. An on-site operator adjudicated success using the physical endpoint criteria in Table~\ref{tab:supp-real-robot-protocol}.

\begin{table}[H]
\centering
\caption{\textbf{Real-robot setup, evaluation protocol, and physical endpoint criteria on the dual-arm G2.} Fig.~\ref{fig:real-robot-sequences} shows representative executions.}
\label{tab:supp-real-robot-protocol}
\renewcommand{\suppstretch}{1.08}\suppstyle
\setlength{\tabcolsep}{6pt}
\begin{tabularx}{\linewidth}{l l Z}
\toprule
\textbf{Item / Task} & \textbf{Interface / Setting} & \textbf{Specification / Physical endpoint} \\
\midrule
Cameras \& perception & Head + dual wrist RGB-D & Head ($640{\times}400$); wrist ($1280{\times}1056$ native for SAM3, calibrated depth back-projection, and VLA; $640{\times}528$ rescaled planner views); calibrated gripper-center TCP kinematics. \\
Reset \& initial layouts & 10 matched layouts & Operator-triggered reset returns arms, grippers, head, and torso to a canonical home pose outside episode timing; ten predefined workspace layouts are matched across $H_0$--$H_3$. \\
Logging \& updates & Offline between batches & Passive three-camera H.264 recording captures rollouts for offline $(C,W,L)$ revision between ten-trial batches while policy weights remain fixed. \\
\midrule
Sorting & Atomic actions & The target object rests inside the designated container after release. \\
Pouring & Frozen VLA (\texttt{vla\_act}) & Water enters the target cup without spillage, and both vessels finish upright on the table. \\
Plugging & Frozen VLA (\texttt{vla\_act}) & The plug is seated in the target socket, and the connected device is placed and released. \\
\bottomrule
\end{tabularx}
\end{table}

%% file: tables/simulator_alignment_states.tex
\centering
\caption{\textbf{State assignments by physical condition.} Unresolved propositions are excluded from precision and recall and retained in the progress and selection bounds.}
\label{tab:alignment-states}
\suppstyle
\begin{tabularx}{\linewidth}{>{\raggedright\arraybackslash}X *{3}{Y}}
\toprule
\suppth{Physical condition} & \suppth{Resolved / total} & \suppth{TP / FP / FN / TN} & \suppth{Precision / recall (\%)} \\
\midrule
Acquisition and control & 18 / 25 & 5 / 0 / 0 / 13 & 100.0 / 100.0 \\
Placement & 22 / 22 & 4 / 0 / 0 / 18 & 100.0 / 100.0 \\
Articulated access & 8 / 10 & 7 / 0 / 0 / 1 & 100.0 / 100.0 \\
\bottomrule
\end{tabularx}

%% file: tables/simulator_alignment_progress.tex
\centering
\caption{\textbf{Progress across the shared-harness optimization.} Every version runs the same 24 task-initialization pairs. Physical progress excludes epistemic milestones. Full progress retains them as unresolved.}
\label{tab:alignment-progress}
\suppstyle
\setlength{\tabcolsep}{4pt}
\begin{tabularx}{\linewidth}{c c c >{\centering\arraybackslash}p{5.2em} Y >{\centering\arraybackslash}p{4.8em} Y}
\toprule
\suppth{Harness} & \suppth{Failures} & \suppth{SR (\%)} & \suppth{Scorer, phys.} & \suppth{Reference, phys.} & \suppth{Scorer, full} & \suppth{Reference, full} \\
\midrule
$H_0$ & 12/24 & 50.00 & 0.6250 & $[0.5903, 0.6250]$ & 0.6979 & $[0.5799, 0.6979]$ \\
$H_1$ & 9/24 & 62.50 & 0.6944 & $[0.6944, 0.6944]$ & 0.7500 & $[0.6944, 0.7500]$ \\
$H_2$ & 3/24 & 87.50 & 0.8889 & $[0.8889, 0.8889]$ & 0.9167 & $[0.8889, 0.9167]$ \\
$H_3$ & 8/24 & 66.67 & 0.7986 & $[0.7361, 0.7986]$ & 0.8125 & $[0.7361, 0.8125]$ \\
$H_4$ & 4/24 & 83.33 & 0.8750 & $[0.8333, 0.8750]$ & 0.9132 & $[0.8333, 0.9132]$ \\
\rowcolor{tableaccent}
$H_5$ & 3/24 & 87.50 & 0.9167 & $[0.8750, 0.9167]$ & 0.9306 & $[0.8750, 0.9306]$ \\
\bottomrule
\end{tabularx}

%% file: tables/simulator_alignment_selection.tex
\centering
\caption{\textbf{Checkpoint selection.} Each row selects from the harnesses up to that version. The reference loss replaces progress in Eq.~\ref{eq:shared-distillation} and keeps official success and code cost. Earlier prefixes apply the same selection rule retrospectively.}
\label{tab:alignment-selection}
\suppstyle
\setlength{\tabcolsep}{4pt}
\begin{tabularx}{\linewidth}{c c c *{3}{Y}}
\toprule
\suppth{Available pool} & \suppth{Selected} & \suppth{$\mathcal L_{\mathrm{meta}}$} & \suppth{Reference loss} & \suppth{Regret, phys.} & \suppth{Regret, full} \\
\midrule
$H_0$-$H_0$ & $H_0$ & 0.8021 & $[0.8021, 0.9201]$ & $[0.0000, 0.0000]$ & $[0.0000, 0.0000]$ \\
$H_0$-$H_1$ & $H_1$ & 0.6518 & $[0.6518, 0.7074]$ & $[0.0000, 0.0000]$ & $[0.0000, 0.0000]$ \\
$H_0$-$H_2$ & $H_2$ & 0.2406 & $[0.2406, 0.2684]$ & $[0.0000, 0.0000]$ & $[0.0000, 0.0000]$ \\
$H_0$-$H_3$ & $H_2$ & 0.2406 & $[0.2406, 0.2684]$ & $[0.0000, 0.0000]$ & $[0.0000, 0.0000]$ \\
$H_0$-$H_4$ & $H_2$ & 0.2406 & $[0.2406, 0.2684]$ & $[0.0000, 0.0000]$ & $[0.0000, 0.0000]$ \\
\rowcolor{tableaccent}
$H_0$-$H_5$ & $H_5$ & 0.2375 & $[0.2375, 0.2930]$ & $[0.0000, 0.0246]$ & $[0.0000, 0.0524]$ \\
\bottomrule
\end{tabularx}

%% file: tables/libero_per_task.tex
\begingroup
\small
\setlength{\aboverulesep}{0pt}\setlength{\belowrulesep}{0pt}
\renewcommand{\arraystretch}{1.08}
\setlength{\tabcolsep}{3pt}
\setlength{\LTcapwidth}{\linewidth}
\setlength{\LTpre}{6pt}\setlength{\LTpost}{6pt}
\begin{longtable}{>{\centering\arraybackslash}p{26pt} *{12}{>{\centering\arraybackslash}p{\dimexpr(\linewidth-26pt-26\tabcolsep)/12\relax}}}
\caption{\textbf{Per-task LIBERO-PRO results.} P/T denote position/task perturbations. Dev. counts mutation rounds. Dec., Out., and Steps average controller turns, output tokens (k), and physical simulator steps over ten test seeds. Time (s) averages wall-clock time, excluding 2 of 800 rollouts with logged API transport or capacity errors on requests lasting at least 30~s.}\label{tab:supp-libero-per-task}\\
\toprule
\multirow{2}{*}{\textbf{Task}} & \multicolumn{6}{c}{\textbf{Position (P)}} & \multicolumn{6}{c}{\textbf{Task (T)}} \\
\cmidrule(lr){2-7}\cmidrule(lr){8-13}
 & \textbf{SR (\%)} & \textbf{Dev.} & \textbf{Dec.} & \textbf{Out. (k)} & \textbf{Steps} & \textbf{Time (s)} & \textbf{SR (\%)} & \textbf{Dev.} & \textbf{Dec.} & \textbf{Out. (k)} & \textbf{Steps} & \textbf{Time (s)} \\
\midrule
\endfirsthead
\multicolumn{13}{@{}l}{\footnotesize\textbf{Table \thetable{} (continued)}} \\
\toprule
\multirow{2}{*}{\textbf{Task}} & \multicolumn{6}{c}{\textbf{Position (P)}} & \multicolumn{6}{c}{\textbf{Task (T)}} \\
\cmidrule(lr){2-7}\cmidrule(lr){8-13}
 & \textbf{SR (\%)} & \textbf{Dev.} & \textbf{Dec.} & \textbf{Out. (k)} & \textbf{Steps} & \textbf{Time (s)} & \textbf{SR (\%)} & \textbf{Dev.} & \textbf{Dec.} & \textbf{Out. (k)} & \textbf{Steps} & \textbf{Time (s)} \\
\midrule
\noalign{\vspace{1.5pt}}
\endhead
\bottomrule
\endfoot
\rowcolor{tablegray}\multicolumn{13}{l}{\textbf{Spatial}} \\*
00 & 100.0 & 2 & 6.1 & 1.26 & 119.5 & 198 & 100.0 & 0 & 12.3 & 3.66 & 638.1 & 593 \\
01 & 100.0 & 5 & 10.9 & 3.01 & 299.8 & 489 & 100.0 & 0 & 6.2 & 1.07 & 241.3 & 329 \\
02 & 100.0 & 0 & 7.0 & 1.33 & 176.0 & 271 & 100.0 & 2 & 8.9 & 1.91 & 208.3 & 316 \\
03 & 90.0 & 4 & 12.8 & 2.66 & 344.1 & 364 & 100.0 & 2 & 13.8 & 2.63 & 269.0 & 380 \\
04 & 100.0 & 0 & 23.8 & 5.86 & 1088.1 & 1190 & 100.0 & 5 & 12.0 & 3.70 & 403.2 & 568 \\
05 & 100.0 & 1 & 7.5 & 1.42 & 300.7 & 339 & 100.0 & 1 & 9.5 & 2.53 & 332.6 & 430 \\
06 & 100.0 & 2 & 13.1 & 2.82 & 453.1 & 684 & 100.0 & 0 & 9.0 & 1.81 & 355.4 & 544 \\
07 & 100.0 & 1 & 8.4 & 1.65 & 322.8 & 369 & 100.0 & 3 & 11.4 & 2.18 & 477.7 & 557 \\
08 & 100.0 & 1 & 9.6 & 2.15 & 261.2 & 283 & 100.0 & 4 & 9.8 & 2.47 & 212.5 & 263 \\
09 & 100.0 & 3 & 14.5 & 2.99 & 477.5 & 489 & 100.0 & 1 & 8.0 & 1.81 & 330.9 & 321 \\
\rowcolor{tablegray}\multicolumn{13}{l}{\textbf{Object}} \\*
00 & 100.0 & 0 & 3.4 & 0.56 & 153.0 & 143 & 100.0 & 0 & 4.5 & 0.67 & 147.5 & 141 \\
01 & 100.0 & 1 & 4.1 & 0.86 & 173.0 & 247 & 100.0 & 0 & 6.3 & 1.23 & 299.0 & 374 \\
02 & 100.0 & 1 & 6.1 & 1.16 & 213.5 & 277 & 100.0 & 0 & 3.4 & 0.46 & 152.4 & 175 \\
03 & 100.0 & 4 & 7.2 & 1.33 & 152.6 & 225 & 100.0 & 2 & 6.7 & 1.35 & 190.9 & 238 \\
04 & 100.0 & 1 & 7.8 & 1.60 & 306.6 & 391 & 100.0 & 0 & 6.4 & 1.14 & 231.8 & 294 \\
05 & 100.0 & 0 & 6.0 & 1.44 & 362.6 & 390 & 100.0 & 4 & 5.9 & 1.52 & 126.5 & 229 \\
06 & 100.0 & 1 & 5.9 & 1.08 & 227.8 & 471 & 100.0 & 0 & 5.5 & 1.03 & 151.8 & 224 \\
07 & 100.0 & 0 & 5.9 & 1.00 & 218.1 & 271 & 100.0 & 2 & 6.5 & 1.56 & 290.6 & 330 \\
08 & 100.0 & 2 & 4.8 & 0.92 & 180.5 & 190 & 100.0 & 1 & 6.7 & 1.63 & 175.2 & 202 \\
09 & 100.0 & 0 & 5.2 & 0.71 & 153.6 & 178 & 100.0 & 0 & 4.8 & 0.66 & 140.6 & 170 \\
\rowcolor{tablegray}\multicolumn{13}{l}{\textbf{Goal}} \\*
00 & 100.0 & 4 & 9.8 & 2.18 & 296.2 & 276 & 80.0 & 6 & 19.4 & 4.85 & 888.6 & 863 \\
01 & 100.0 & 1 & 10.6 & 2.51 & 291.3 & 414 & 100.0 & 0 & 6.1 & 1.02 & 228.6 & 282 \\
02 & 100.0 & 0 & 8.2 & 1.72 & 470.3 & 482 & 100.0 & 1 & 17.9 & 3.92 & 737.6 & 788 \\
03 & 100.0 & 4 & 6.7 & 1.10 & 378.5 & 313 & 100.0 & 5 & 9.7 & 2.10 & 453.3 & 394 \\
04 & 100.0 & 0 & 5.9 & 1.13 & 264.3 & 324 & 100.0 & 1 & 7.2 & 1.21 & 326.5 & 363 \\
05 & 100.0 & 6 & 33.6 & 8.39 & 1151.5 & 1326 & 100.0 & 3 & 20.6 & 5.02 & 643.4 & 738 \\
06 & 100.0 & 3 & 9.2 & 2.84 & 290.1 & 372 & 100.0 & 1 & 13.7 & 2.78 & 274.5 & 398 \\
07 & 100.0 & 0 & 2.0 & 0.11 & 83.3 & 110 & 100.0 & 0 & 3.0 & 0.35 & 159.3 & 167 \\
08 & 100.0 & 2 & 14.0 & 3.22 & 521.0 & 479 & 100.0 & 2 & 10.5 & 2.53 & 414.0 & 402 \\
09 & 100.0 & 1 & 14.3 & 2.83 & 643.9 & 584 & 100.0 & 0 & 5.9 & 1.27 & 336.6 & 345 \\
\rowcolor{tablegray}\multicolumn{13}{l}{\textbf{Long}} \\*
00 & 100.0 & 6 & 14.5 & 3.95 & 457.6 & 701 & 100.0 & 6 & 14.0 & 3.16 & 447.7 & 634 \\
01 & 90.0 & 6 & 14.8 & 3.35 & 302.8 & 457 & 100.0 & 1 & 10.4 & 2.71 & 607.0 & 687 \\
02 & 100.0 & 4 & 9.9 & 1.95 & 369.6 & 421 & 100.0 & 5 & 5.2 & 0.75 & 336.1 & 295 \\
03 & 100.0 & 6 & 17.5 & 4.60 & 779.6 & 745 & 70.0 & 10 & 33.4 & 7.46 & 1247.0 & 1173 \\
04 & 100.0 & 7 & 12.2 & 3.31 & 373.3 & 509 & 100.0 & 6 & 18.5 & 5.38 & 490.6 & 655 \\
05 & 100.0 & 6 & 12.5 & 5.34 & 140.9 & 418 & 100.0 & 3 & 12.2 & 5.12 & 475.1 & 626 \\
06 & 90.0 & 6 & 13.1 & 3.42 & 713.0 & 720 & 100.0 & 7 & 15.5 & 4.58 & 585.2 & 749 \\
07 & 100.0 & 6 & 14.6 & 5.28 & 442.7 & 625 & 100.0 & 0 & 15.1 & 3.64 & 597.0 & 615 \\
08 & 100.0 & 6 & 27.5 & 7.49 & 1118.0 & 1050 & 100.0 & 1 & 9.3 & 2.06 & 333.3 & 325 \\
09 & 80.0 & 10 & 42.1 & 8.93 & 1334.5 & 1348 & 100.0 & 6 & 26.4 & 8.91 & 1456.5 & 1130 \\
\end{longtable}
\endgroup

%% file: tables/robodojo_per_task.tex
\begingroup
\small
\setlength{\aboverulesep}{0pt}\setlength{\belowrulesep}{0pt}
\renewcommand{\arraystretch}{0.94}
\setlength{\tabcolsep}{3pt}
\setlength{\LTcapwidth}{\linewidth}
\setlength{\LTpre}{6pt}\setlength{\LTpost}{6pt}
\begin{longtable}{>{\raggedright\arraybackslash}p{.29\linewidth} *{8}{>{\centering\arraybackslash}p{\dimexpr(.71\linewidth-18\tabcolsep)/8\relax}}}
\caption{\textbf{Per-task RoboDojo results.} Generalization pools five standard and five randomized trials. Dev. counts mutation rounds. Dec., Out., and Steps average controller turns, output tokens (k), and physical simulator steps over ten trials. Time (s) averages wall-clock time, excluding 6 of 420 rollouts with logged API transport or capacity errors on requests lasting at least 30~s.}\label{tab:supp-robodojo-per-task}\\
\toprule
\multicolumn{1}{c}{\textbf{Task}} & \textbf{Successes} & \textbf{Score $\uparrow$} & \textbf{SR $\uparrow$} & \textbf{Dev.} & \textbf{Dec. $\downarrow$} & \textbf{Out. (k) $\downarrow$} & \textbf{Steps $\downarrow$} & \textbf{Time (s) $\downarrow$} \\
\midrule
\endfirsthead
\multicolumn{9}{@{}l}{\footnotesize\textbf{Table \thetable{} (continued)}} \\
\toprule
\multicolumn{1}{c}{\textbf{Task}} & \textbf{Successes} & \textbf{Score $\uparrow$} & \textbf{SR $\uparrow$} & \textbf{Dev.} & \textbf{Dec. $\downarrow$} & \textbf{Out. (k) $\downarrow$} & \textbf{Steps $\downarrow$} & \textbf{Time (s) $\downarrow$} \\
\midrule
\noalign{\vspace{1.5pt}}
\endhead
\bottomrule
\endfoot
\midrule
\rowcolor{tableaccent}
\textbf{Five-capability mean} & & \textbf{63.43} & \textbf{58.25} & & \textbf{44.0} & \textbf{21.56} & \textbf{641.2} & \textbf{2129} \\
\bottomrule
\endlastfoot
\rowcolor{tablegray}\multicolumn{9}{l}{\textbf{Generalization}} \\*
Stack bowls & 9/10 & 91.50 & 90.00 & 2 & 43.1 & 21.76 & 593.2 & 1582 \\
Push T & 8/10 & 80.00 & 80.00 & 10 & 70.3 & 26.62 & 430.2 & 6878 \\
Pack objects into box & 0/10 & 7.00 & 0.00 & 10 & 66.3 & 25.38 & 1067.1 & 3691 \\
Fold clothes & 0/10 & 12.00 & 0.00 & 10 & 34.1 & 16.44 & 485.9 & 2023 \\
Hang mugs & 0/10 & 18.50 & 0.00 & 2 & 56.1 & 38.53 & 746.5 & 2675 \\
Sweep blocks & 0/10 & 0.00 & 0.00 & 10 & 64.3 & 24.97 & 896.9 & 2873 \\
Pour liquid into cup & 6/10 & 60.00 & 60.00 & 10 & 41.0 & 17.36 & 334.5 & 2342 \\
Make toast & 4/10 & 50.00 & 40.00 & 10 & 60.1 & 44.21 & 1026.5 & 2901 \\
Arrange largest number & 8/10 & 86.50 & 80.00 & 6 & 64.4 & 16.37 & 875.0 & 1837 \\
Sort nesting dolls by size & 5/10 & 50.00 & 50.00 & 4 & 47.4 & 17.88 & 904.6 & 1939 \\
Store laptop and headphones & 0/10 & 0.00 & 0.00 & 10 & 50.0 & 20.74 & 688.4 & 2004 \\
Stack blocks & 10/10 & 100.00 & 100.00 & 2 & 30.5 & 8.39 & 418.0 & 2138 \\
\rowcolor{tablegray}\multicolumn{9}{l}{\textbf{Precision}} \\*
Fasten screws & 10/10 & 100.00 & 100.00 & 6 & 75.2 & 21.02 & 1181.0 & 1599 \\
Insert tubes & 0/10 & 8.00 & 0.00 & 10 & 60.2 & 26.88 & 469.3 & 3290 \\
Plug in charger & 6/10 & 60.00 & 60.00 & 10 & 35.5 & 21.64 & 353.5 & 1568 \\
Pour balls into vase & 5/10 & 50.00 & 50.00 & 1 & 36.8 & 27.77 & 499.7 & 1623 \\
Play Xylophone & 10/10 & 100.00 & 100.00 & 6 & 41.7 & 18.50 & 408.4 & 1429 \\
Deposit coin & 3/10 & 40.00 & 30.00 & 9 & 30.7 & 19.77 & 272.1 & 1398 \\
Insert key & 0/10 & 13.50 & 0.00 & 10 & 32.1 & 15.03 & 290.0 & 1302 \\
Build tower & 0/10 & 17.00 & 0.00 & 10 & 37.9 & 27.57 & 978.0 & 2968 \\
\rowcolor{tablegray}\multicolumn{9}{l}{\textbf{Long-Horizon}} \\*
Put bottles into dustbin & 9/10 & 94.00 & 90.00 & 1 & 51.4 & 30.04 & 627.0 & 1698 \\
Play tic tac toe & 10/10 & 100.00 & 100.00 & 3 & 22.8 & 8.36 & 942.8 & 1152 \\
Classify objects & 7/10 & 79.50 & 70.00 & 1 & 44.7 & 23.33 & 984.9 & 1669 \\
Fill pen holder & 0/10 & 8.00 & 0.00 & 6 & 81.9 & 54.86 & 936.0 & 5264 \\
Fill egg holder & 0/10 & 8.00 & 0.00 & 10 & 51.0 & 19.55 & 621.3 & 2624 \\
Organize table & 2/10 & 72.50 & 20.00 & 10 & 73.3 & 54.34 & 934.2 & 2805 \\
Play stacking toy & 6/10 & 75.00 & 60.00 & 1 & 69.0 & 60.31 & 1122.2 & 3694 \\
Make kong & 0/10 & 0.00 & 0.00 & 10 & 50.0 & 20.68 & 555.8 & 2389 \\
\rowcolor{tablegray}\multicolumn{9}{l}{\textbf{Memory}} \\*
Cover blocks & 10/10 & 100.00 & 100.00 & 1 & 27.0 & 7.66 & 740.1 & 756 \\
Match and pick from conveyor & 8/10 & 80.00 & 80.00 & 6 & 32.3 & 8.79 & 434.8 & 1347 \\
Swap T & 9/10 & 90.00 & 90.00 & 1 & 27.0 & 21.21 & 379.6 & 1121 \\
Press by number & 10/10 & 100.00 & 100.00 & 1 & 28.6 & 6.21 & 483.3 & 2207 \\
Imitate sorting sequence & 10/10 & 100.00 & 100.00 & 8 & 57.1 & 13.65 & 1210.1 & 1935 \\
Swap blocks & 9/10 & 90.00 & 90.00 & 1 & 31.7 & 7.00 & 532.0 & 1448 \\
\rowcolor{tablegray}\multicolumn{9}{l}{\textbf{Open}} \\*
Align blocks & 8/10 & 80.00 & 80.00 & 2 & 14.9 & 6.82 & 152.7 & 1101 \\
Solve equation & 10/10 & 100.00 & 100.00 & 1 & 13.0 & 4.08 & 206.7 & 580 \\
Stack blocks by language & 10/10 & 100.00 & 100.00 & 4 & 33.6 & 7.39 & 364.7 & 2788 \\
General pickup & 10/10 & 100.00 & 100.00 & 1 & 10.6 & 2.76 & 99.8 & 324 \\
Classify objects by language & 8/10 & 85.00 & 80.00 & 1 & 56.3 & 31.81 & 995.8 & 2342 \\*
Pick from conveyor by image & 5/10 & 50.00 & 50.00 & 0 & 47.9 & 27.31 & 511.2 & 2400 \\*
Store tools in toolbox & 0/10 & 12.50 & 0.00 & 10 & 58.7 & 30.42 & 824.2 & 3023 \\*
Pour by language & 6/10 & 67.00 & 60.00 & 10 & 41.2 & 30.47 & 633.5 & 2472 \\*
\end{longtable}
\endgroup

%% file: main.bbl
\begin{thebibliography}{38}
\providecommand{\natexlab}[1]{#1}
\providecommand{\url}[1]{\texttt{#1}}
\expandafter\ifx\csname urlstyle\endcsname\relax
  \providecommand{\doi}[1]{doi: #1}\else
  \providecommand{\doi}{doi: \begingroup \urlstyle{rm}\Url}\fi

\bibitem[Ahn et~al.(2022)Ahn, Brohan, Brown, Chebotar, Cortes, David, Finn, Fu,
  Gopalakrishnan, Hausman, et~al.]{ahn2022can}
Michael Ahn, Anthony Brohan, Noah Brown, Yevgen Chebotar, Omar Cortes, Byron
  David, Chelsea Finn, Chuyuan Fu, Keerthana Gopalakrishnan, Karol Hausman,
  et~al.
\newblock Do as i can, not as i say: Grounding language in robotic affordances.
\newblock \emph{arXiv preprint arXiv:2204.01691}, \unskip\nobreakspace{}2022.

\bibitem[{Awomo-WAM Team}(2026)]{team2026awomo}
{Awomo-WAM Team}.
\newblock Awomo-0.5 technical report.
\newblock
  \href{https://awomo-westlakedi.github.io/Awomo-0.5-page/assets/awomo-0.5-report.pdf}{Technical
  report}, \unskip\nobreakspace{}2026.

\bibitem[Chen et~al.(2026)Chen, Chen, Li, Tang, Su, Lu, Wan, Chen, Liu, Yan,
  et~al.]{chen2026robodojo}
Tianxing Chen, Yue Chen, Zixuan Li, Junyuan Tang, Kailun Su, Haoran Lu, Weijie
  Wan, Baijun Chen, Songling Liu, Haowen Yan, et~al.
\newblock Robodojo: A unified sim-and-real benchmark for comprehensive
  evaluation of generalist robot manipulation policies.
\newblock \emph{arXiv preprint arXiv:2607.04434}, \unskip\nobreakspace{}2026.

\bibitem[Chu et~al.(2026)Chu, Barr, Goues, O'Hearn, Harman, Sarro, and
  Ye]{chu2026openrua}
Zhaoyang Chu, Earl~T Barr, Claire~Le Goues, Peter O'Hearn, Mark Harman,
  Federica Sarro, and He~Ye.
\newblock Openrua: Robot-use agents are zero-shot visuomotor policies.
\newblock \emph{arXiv preprint arXiv:2610.02459}, \unskip\nobreakspace{}2026.

\bibitem[Ding et~al.(2026)Ding, Mi, Huang, Wang, Zhang, Hao, Chen, Li, Zheng,
  Guo, et~al.]{ding2026zetta}
Xin Ding, Liang Mi, Mingzhe Huang, Zixuan Wang, Chao Zhang, Zixu Hao, Fu~Chen,
  Xiangyu Li, Yikai Zheng, Yaoyu Guo, et~al.
\newblock Zetta $\zeta$: An efficient closed-loop embodied harness for
  self-evolving physical intelligence.
\newblock \emph{arXiv preprint arXiv:2608.16590}, \unskip\nobreakspace{}2026.

\bibitem[Du et~al.(2026)Du, Guo, Zhang, Wang, Dong, Gao, Han, Lu, Yu, Shen,
  et~al.]{du2026vlm}
Tingwen Du, Haoyu Guo, Hanxue Zhang, Xuanxu Wang, Junting Dong, Chao Gao,
  Mingfei Han, Tao Lu, Mulin Yu, Chunhua Shen, et~al.
\newblock From vlm planners to robotic agents: Self-evolving harnesses for
  history-dependent manipulation.
\newblock \emph{arXiv preprint arXiv:2610.29560}, \unskip\nobreakspace{}2026.

\bibitem[Elmaaroufi et~al.(2026)Elmaaroufi, Svegliato, Kalade, Schelle, Seshia,
  and Zaharia]{elmaaroufi2026rho}
Karim Elmaaroufi, Justin Svegliato, Sarunas Kalade, Graham Schelle, Sanjit~A
  Seshia, and Matei Zaharia.
\newblock Rho: Your coding agent is secretly a roboticist.
\newblock \emph{arXiv preprint arXiv:2606.16458}, \unskip\nobreakspace{}2026.

\bibitem[Gao et~al.(2026{\natexlab{a}})Gao, Zhou, Zhu, Zheng, Ge, Li, Wu, Wang,
  Zeng, Shangguan, et~al.]{gao2026hexashell}
Hongcheng Gao, Alex Zhou, Jay Zhu, Zelin Zheng, Shijia Ge, Xiping Li, Di~Wu,
  Yazhe Wang, Jianshu Zeng, Xuan Shangguan, et~al.
\newblock Hexashell: A thin harness for exploring the limits of vlm agents as
  robot coding agents.
\newblock \emph{arXiv preprint arXiv:2610.23479},
  \unskip\nobreakspace{}2026{\natexlab{a}}.

\bibitem[Gao et~al.(2026{\natexlab{b}})Gao, Zhou, Zheng, Ge, Zhu, Wang, Zeng,
  Shangguan, Wu, He, Jia, Wu, and He]{gao2026selfevolving}
Hongcheng Gao, Jingjing Zhou, Zelin Zheng, Shijia Ge, Jay Zhu, Yazhe Wang,
  Jianshu Zeng, Xuan Shangguan, Di~Wu, Lingyu He, Zhiqi Jia, Sihang Wu, and
  Xiao He.
\newblock Self-evolving coding agents: From digital programs to physical-world
  intelligence.
\newblock \emph{arXiv preprint arXiv:2609.35432},
  \unskip\nobreakspace{}2026{\natexlab{b}}.

\bibitem[Gu et~al.(2026)Gu, Zhu, Lin, Zhu, Chen, Wang, Yuan, Yang, Zhao, Li,
  et~al.]{gu2026harnesswam}
Zhaopeng Gu, Bingke Zhu, Tianxi Lin, Guibo Zhu, Yingying Chen, Kai Wang, Tingyu
  Yuan, Fan Yang, Chaoyang Zhao, Zhaowen Li, et~al.
\newblock Harnesswam: Bridging prediction and deliberation in world action
  models.
\newblock \emph{arXiv preprint arXiv:2608.09516}, \unskip\nobreakspace{}2026.

\bibitem[Guo et~al.(2026)Guo, Yan, Zhong, Zhang, Yang, Lu, Chen, Wang, Deng,
  Yang, et~al.]{guo2026video}
Yanjiang Guo, Haodong Yan, Zhide Zhong, Zhongru Zhang, Qingyuan Yang, Qingzhou
  Lu, Xiaoyu Chen, Yen-Jen Wang, Shuying Deng, Chenghan Yang, et~al.
\newblock Video prediction policy 2: Predict better, act better.
\newblock \emph{arXiv preprint arXiv:2610.10270}, \unskip\nobreakspace{}2026.

\bibitem[Han et~al.(2024)Han, Zhu, Zhu, Wu, and Zhu]{han2024interpret}
Muzhi Han, Yifeng Zhu, Song-Chun Zhu, Ying~Nian Wu, and Yuke Zhu.
\newblock Interpret: Interactive predicate learning from language feedback for
  generalizable task planning.
\newblock \emph{arXiv preprint arXiv:2405.19758}, \unskip\nobreakspace{}2024.

\bibitem[Hu et~al.(2026)Hu, Hu, Hong, Chen, and Liu]{hu2026embodied}
Kairui Hu, Siyuan Hu, Fangzhou Hong, Zhaoxi Chen, and Ziwei Liu.
\newblock Embodied turing machines: Stateful code for robot recursive
  self-improvement.
\newblock \emph{arXiv preprint arXiv:2610.12369}, \unskip\nobreakspace{}2026.

\bibitem[Huang et~al.(2026)Huang, Li, Hu, Qi, Xiao, Coates, Cao, Zhang, and
  Zhang]{huang2026roboharness}
Jinbang Huang, Zhiyuan Li, Yuanzhao Hu, Ran Qi, Yixin Xiao, Mark Coates,
  Tongtong Cao, Zhanguang Zhang, and Yingxue Zhang.
\newblock Roboharness: Memory-driven orchestration of heterogeneous robot
  policies for long-horizon planning.
\newblock \emph{arXiv preprint arXiv:2607.18060}, \unskip\nobreakspace{}2026.

\bibitem[Huang et~al.(2022)Huang, Xia, Xiao, Chan, Liang, Florence, Zeng,
  Tompson, Mordatch, Chebotar, et~al.]{huang2022inner}
Wenlong Huang, Fei Xia, Ted Xiao, Harris Chan, Jacky Liang, Pete Florence, Andy
  Zeng, Jonathan Tompson, Igor Mordatch, Yevgen Chebotar, et~al.
\newblock Inner monologue: Embodied reasoning through planning with language
  models.
\newblock \emph{arXiv preprint arXiv:2207.05608}, \unskip\nobreakspace{}2022.

\bibitem[Intelligence et~al.(2025)Intelligence, Black, Brown, Darpinian,
  Dhabalia, Driess, Esmail, Equi, Finn, Fusai, et~al.]{intelligence2025pi_}
Physical Intelligence, Kevin Black, Noah Brown, James Darpinian, Karan
  Dhabalia, Danny Driess, Adnan Esmail, Michael Equi, Chelsea Finn, Niccolo
  Fusai, et~al.
\newblock $\pi_{0.5}$: a vision-language-action model with open-world
  generalization.
\newblock \emph{arXiv preprint arXiv:2504.16054}, \unskip\nobreakspace{}2025.

\bibitem[Kim et~al.(2026)Kim, Lee, Kim, Kim, and Kwak]{kim2026recursive}
Seungyeon Kim, Junhoo Lee, Minkyu Kim, Baekseung Kim, and Nojun Kwak.
\newblock Recursive harness distillation across agents for robot manipulation.
\newblock \emph{arXiv preprint arXiv:2609.33378}, \unskip\nobreakspace{}2026.

\bibitem[Li et~al.(2026{\natexlab{a}})Li, Zhou, Zhu, Chen, Wang, Hu, Yu, Jiang,
  Su, Ma, et~al.]{li2026roboclaw}
Ruiying Li, Yunlang Zhou, YuYao Zhu, Kylin Chen, Sukai Wang, Kongtao Hu, Minhui
  Yu, Bowen Jiang, Zhan Su, Jiayao Ma, et~al.
\newblock Roboclaw: An agentic framework for scalable long-horizon robotic
  tasks.
\newblock In \emph{European conference on computer vision},
  \unskip\nobreakspace{}2026{\natexlab{a}}.

\bibitem[Li et~al.(2026{\natexlab{b}})Li, Chi, Zheng, Liu, Wang, Gao, Jiang,
  Hong, Xiao, Yang, et~al.]{li2026embodiedevo}
Wenyi Li, Yizhe Chi, Youjie Zheng, Chunyu Liu, Xiaoqiu Wang, Mingju Gao, Zheng
  Jiang, Deyao Hong, Calvin Xiao, Kaisen Yang, et~al.
\newblock Embodiedevo: Self-evolving coding agent for physical intelligence.
\newblock \emph{arXiv preprint arXiv:2610.18867},
  \unskip\nobreakspace{}2026{\natexlab{b}}.

\bibitem[Liang et~al.(2023)Liang, Huang, Xia, Xu, Hausman, Ichter, Florence,
  and Zeng]{liang2023code}
Jacky Liang, Wenlong Huang, Fei Xia, Peng Xu, Karol Hausman, Brian Ichter, Pete
  Florence, and Andy Zeng.
\newblock Code as policies: Language model programs for embodied control.
\newblock In \emph{2023 IEEE International conference on robotics and
  automation (ICRA)}, \unskip\nobreakspace{}2023.

\bibitem[Liu et~al.(2026)Liu, Shi, Sang, He, Lin, Wei, Wang, Dumoulin, Jin, and
  Lu]{liu2026adaptive}
Zewen Liu, Zhan Shi, Yisi Sang, Bing He, Minhua Lin, Tianxin Wei, Dakuo Wang,
  Benoit Dumoulin, Wei Jin, and Hanqing Lu.
\newblock Adaptive auto-harness: Sustained self-improvement for agentic system
  deployment on open-ended task streams.
\newblock \emph{arXiv preprint arXiv:2606.01770}, \unskip\nobreakspace{}2026.

\bibitem[Lu et~al.(2026)Lu, Wu, Kou, Lin, Fu, Xiao, Mandlekar, Xu, Shi,
  Goldberg, Chen, Chowdhury, Kautz, Zhu, Fan, and Wang]{lu2026aspire}
Runyu Lu, Yubo Wu, Ethan Kou, Haotian Lin, Letian Fu, Wenli Xiao, Ajay
  Mandlekar, Yinzhen Xu, Guanya Shi, Ken Goldberg, Ang Chen, Mosharaf
  Chowdhury, Jan Kautz, Yuke Zhu, Linxi~"Jim" Fan, and Guanzhi Wang.
\newblock Aspire: Agentic /skills discovery for robotics.
\newblock \emph{arXiv preprint arXiv:2607.00272}, \unskip\nobreakspace{}2026.

\bibitem[{Manifold AI Team}(2026)]{team2026r}
{Manifold AI Team}.
\newblock {R$^2$SI}: Recursive robot self-improvement via a bi-level agentic
  system with agent--model co-evolution.
\newblock \href{https://r2si.worldscape.io/report.pdf}{Technical report},
  \unskip\nobreakspace{}2026.

\bibitem[{MMLab \& PhysicalRSI contributors}(2026)]{physicalrsi2026}
{MMLab \& PhysicalRSI contributors}.
\newblock {Physical RSI 1.0: Recursive Self-Harness for Scaling Embodied
  Skills}.
\newblock \url{https://mmlab.hk/research/PhysicalRSI},
  \unskip\nobreakspace{}2026.

\bibitem[Rana et~al.(2023)Rana, Haviland, Garg, Abou-Chakra, Reid, and
  Suenderhauf]{rana2023sayplan}
Krishan Rana, Jesse Haviland, Sourav Garg, Jad Abou-Chakra, Ian Reid, and Niko
  Suenderhauf.
\newblock Sayplan: Grounding large language models using 3d scene graphs for
  scalable robot task planning.
\newblock \emph{arXiv preprint arXiv:2307.06135}, \unskip\nobreakspace{}2023.

\bibitem[Sun et~al.(2026)Sun, Xu, Cao, Liu, Sun, Pang, Zhang, Yang, Pang, He,
  et~al.]{sun2026atomvla}
Xiaoquan Sun, Zetian Xu, Chen Cao, Zonghe Liu, Yihan Sun, Jingrui Pang, Ruijian
  Zhang, Zhen Yang, Kang Pang, Dingxin He, et~al.
\newblock Atomvla: Scalable post-training for robotic manipulation via
  predictive latent world models.
\newblock \emph{arXiv preprint arXiv:2603.08519}, \unskip\nobreakspace{}2026.

\bibitem[Wang et~al.(2026{\natexlab{a}})Wang, Chang, Ma, Yang, Feng, Liu, Yang,
  Cao, Wang, Su, and Wang]{wang2026adarsi}
Peidong Wang, Ying Chang, Zhiming Ma, Xiaocui Yang, Shi Feng, Daqi Liu, Yuzhe
  Yang, Xu~Cao, Chuang Wang, Jian Su, and Daling Wang.
\newblock {AdaRSI}: On-demand recursive self-improvement for embodied agents.
\newblock \emph{arXiv preprint arXiv:2610.20474},
  \unskip\nobreakspace{}2026{\natexlab{a}}.

\bibitem[Wang et~al.(2026{\natexlab{b}})Wang, Ma, Chang, Luo, Zhang, Wang,
  Yang, Feng, Yang, and Li]{wang2026self}
Peidong Wang, Zhiming Ma, Ying Chang, Xufang Luo, Yiqun Zhang, Zihan Wang,
  Xiaocui Yang, Shi Feng, Yuqing Yang, and Dongsheng Li.
\newblock Self-evolving embodied agents via skill-harness evolution.
\newblock \emph{arXiv preprint arXiv:2608.11350},
  \unskip\nobreakspace{}2026{\natexlab{b}}.

\bibitem[Wang et~al.(2026{\natexlab{c}})Wang, Wu, Zhang, Wang, Shao, Luan, Li,
  Li, Gu, Zhou, et~al.]{wang2026harnesspai}
Xin Wang, Wenhao Wu, Menghao Zhang, Zhi Wang, Kun Shao, Jian Luan, Yang Li,
  Qing Li, Shangding Gu, Huichi Zhou, et~al.
\newblock Harnesspai: An evolving harness for physical ai.
\newblock \emph{arXiv preprint arXiv:2609.29166},
  \unskip\nobreakspace{}2026{\natexlab{c}}.

\bibitem[Xiao et~al.(2026)Xiao, Xie, Zhang, Lin, Fu, Xue, Lu, Yang, Dai, Wang,
  et~al.]{xiao2026enpire}
Wenli Xiao, Jia Xie, Tonghe Zhang, Haotian Lin, Letian Fu, Haoru Xue, Jalen Lu,
  Yi~Yang, Cunxi Dai, Zi~Wang, et~al.
\newblock Enpire: Agentic robot policy self-improvement in the real world.
\newblock \emph{arXiv preprint arXiv:2606.19980}, \unskip\nobreakspace{}2026.

\bibitem[Xu et~al.(2026{\natexlab{a}})Xu, Xu, Chen, Dong, Zhang, Xue, Liu, and
  Zhang]{xu2026tarsphere}
Yicheng Xu, Zhikai Xu, Ziyuan Chen, Yuhang Dong, Tao Zhang, Zhucun Xue, Yong
  Liu, and Jiangning Zhang.
\newblock Tarsphere: A robot-oriented full-stack framework for long-horizon
  embodied agents in evolving physical and social environments.
\newblock \emph{arXiv preprint arXiv:2610.30043},
  \unskip\nobreakspace{}2026{\natexlab{a}}.

\bibitem[Xu et~al.(2026{\natexlab{b}})Xu, Zhang, Xue, Hu, Yi, Hu, Liu, and
  Tao]{xu2026uniicl}
Yicheng Xu, Jiangning Zhang, Zhucun Xue, Teng Hu, Ran Yi, Xiaobin Hu, Yong Liu,
  and Dacheng Tao.
\newblock {UniICL}: Systematizing unified multimodal in-context learning
  through a capability-oriented taxonomy.
\newblock \emph{arXiv preprint arXiv:2603.24690},
  \unskip\nobreakspace{}2026{\natexlab{b}}.

\bibitem[Yang et~al.(2026)Yang, Chen, Gao, and Zhang]{ora2026code}
Yuanbo Yang, Siyi Chen, Jun Gao, and Hao Zhang.
\newblock {ORA}: Open robot-use agents, \unskip\nobreakspace{}October 2026.
\newblock \mbox{URL \url{https://github.com/hao-ai-lab/ORA}}.
\newblock Blog post: https://haoailab.com/blogs/ora/.

\bibitem[Yu et~al.(2026)Yu, Li, Li, Yan, Peng, Zhong, Xin, Yao, Tian, and
  Li]{yu2026robodoctor}
Heyang Yu, Haomin Li, Sihang Li, Jixin Yan, Shiyun Peng, Wanqi Zhong, Tianyu
  Xin, Yuan Yao, Xiaoyu Tian, and Yiming Li.
\newblock Robodoctor: Training-free recursive self-improvement for robot
  policies., \unskip\nobreakspace{}2026.

\bibitem[Zang et~al.(2025)Zang, Wei, Xu, Wu, Guo, Wang, Lin, Wang, Shi, Xie,
  et~al.]{zang2025rlinf}
Hongzhi Zang, Mingjie Wei, Si~Xu, Yongji Wu, Zhen Guo, Yuanqing Wang, Hao Lin,
  Peihong Wang, Liangzhi Shi, Yuqing Xie, et~al.
\newblock Rlinf-vla: A unified and efficient framework for reinforcement
  learning of vision-language-action models.
\newblock \emph{arXiv preprint arXiv:2510.06710}, \unskip\nobreakspace{}2025.

\bibitem[Zhang et~al.(2026)Zhang, Zhang, Gao, Li, Liu, Nie, Zhu, Qiu, Yan, Liu,
  et~al.]{zhang2026harness}
Yixian Zhang, Huanming Zhang, Feng Gao, Xiao Li, Zhihao Liu, Yi~Nie, Chunyang
  Zhu, Jiaxing Qiu, Yuchen Yan, Jiyuan Liu, et~al.
\newblock Harness vla: Steering frozen vlas into reliable manipulation
  primitives via memory-guided agents.
\newblock \emph{arXiv preprint arXiv:2607.08448}, \unskip\nobreakspace{}2026.

\bibitem[Zheng et~al.(2026)Zheng, Li, Wang, Liu, Kang, Feng, Zheng, Zou, Chen,
  Zeng, et~al.]{zheng2026x}
Jinliang Zheng, Jianxiong Li, Zhihao Wang, Dongxiu Liu, Xirui Kang, Yuchun
  Feng, Yinan Zheng, Jiayin Zou, Yilun Chen, Jia Zeng, et~al.
\newblock X-vla: Soft-prompted transformer as scalable cross-embodiment
  vision-language-action model.
\newblock In \emph{International Conference on Learning Representations},
  \unskip\nobreakspace{}2026.

\bibitem[Zhou et~al.(2025)Zhou, Xu, Tie, Chen, Zhang, Chu, Zhou, and
  Sun]{zhou2025libero}
Xueyang Zhou, Yangming Xu, Guiyao Tie, Yongchao Chen, Guowen Zhang, Duanfeng
  Chu, Pan Zhou, and Lichao Sun.
\newblock Libero-pro: Towards robust and fair evaluation of
  vision-language-action models beyond memorization.
\newblock \emph{arXiv preprint arXiv:2510.03827}, \unskip\nobreakspace{}2025.

\end{thebibliography}
